MANCHESTER
1824
The University of Manchester

# The Design and Simulation of Biomimetic Fish Robot for Aquatic Creature Study

Third Year Individual Project – Final Report

April 2019

Ningzhe Hou

10407299

Supervisor: Dr Alexandru Stancu

School of Electrical and Electronic Engineering

# Contents

Total word count: 8730

# List of Figure

# List of Table

**Abstract**

In the application of underwater creature study, comparing with propeller-powered ROVs and servo motor actuated robotic fish, novel biomimetic fish robot designs with soft actuation structure could interact with aquatic creatures closely and record authentic habitats and behaviours.

This final project report presents the detailed design process of a hydraulic soft actuator powered robotic fish for aquatic creature study capable of swimming along the 3D trajectory. The robotic fish is designed based on the analysis of the pro and cons of existing designs.

Except for the mechanical and electronic designs and manufacturing method of crucial components, a simplified open-loop control algorithm was designed to check the functionality of the application board and microcontroller in the Proteus simulation environment. As the key component of the robotic fish, Finite Element Method (FEM) simulations were conducted to visualise the soft actuator's deformation under different pressure to validate the design. Computational Fluid Dynamics (CFD) simulations were also conducted to improve the hydrodynamic efficiency of the shape of robotic fish.

Although physical manufacturing is impossible due to the pandemic, the simulations show overall good performance in terms of control, actuation, and hydrodynamic efficiency.

# Acknowledgement

Although it has been a special and challenging year due to the outbreak of the pandemic, I have received many supports throughout the final project period.

I would first like to express my most tremendous gratitude to my project supervisor, Dr Alexandru Stancu, who agreed to supervise the bespoke project and provided me with valuable technical and executive support throughout the project. Without him, I would not be able to conduct research on such a topic that I have a great passion for at the beginning. The technical knowledge and resources he provided are of great importance to the implementation of the project as well.

I also want to express my gratitude to Dr Alexandru Stancu's colleagues, Mario Guillermo Martínez Guerrero and Zachary Madin, who helped me with the research on possible 3D printing method and material of soft actuation structure.

I would also like to thank Professor Guangming Xie and Dr Xingwen Zheng from my internship at Peking University Intelligent Biomimetic Design Lab. They led me into the field of soft robotic and cultivated my academic research capacities and attitude.

In addition, I want to acknowledge with gratitude for the love, caring, and financial supports from my parents, Furong Tang and Wanfu Hou. They did not only keep encouraging me during the most depressing days. I would not be able to fully focus on my project without worrying about the living cost without their support.

# 1. Introduction

## 1.1. Background

Remotely Operated Underwater Vehicle (ROV) is widely used for underwater life study nowadays due to their high capacity to carry sampling and operating equipment and high mobility. However, the main functionality, even for large work-class ROVs, is taking pictures of the worksite. [1] In the case of underwater creature study, the ROVs with loud noise generated by propeller and large size would disturb the objects of research and make them instinctively escape from the ROVs, which is non-ideal for the study on the natural habitats of the underwater creatures as well as the protection of wildlife. A biomimetic robotic fish with a camera could be a perfect substitution for the ROVs in applying underwater creature study, as they could be integrated into the natural environment without disturbing the objects of study and observing authentic habitats and behaviours.

With centuries of evolution, aquatic creatures have developed efficient swimming mechanisms. The swimming efficiency of fish could reach up to 90 % efficiency, while most conventional human-made propellers could only reach 40 to 50 % efficiency.[2] The researchers created various low-noise and high-efficient Autonomous Underwater Vehicles by studying the fishes' swimming mechanism.

In the design of biomimetic fish robots, most of the designs still use servo motors and multi-link structure as the tail fin and lateral fin actuator. The robot fish with such design could not wholly imitate fishes' behaviour due to the limitation on the degrees of freedom, and the control algorithm is also complicated. The control of traditional robotic fish actuators with oscillatory hinge joints normally needs to establish a general kinematical model and develop control methods based on the model. [2] Moreover, the complexity will significantly go up when more hinge joints are designed. The overall system power consumption will significantly increase when more servo motors are used to increase the degree of freedom of the tail fin actuator.

Soft robotics also has excellent potential in the application of biomimetic designs since the actuation structure and compliant materials are widely used in soft robotics design, and they share similarities with the organic structures on the creatures. The research and application of soft robotics have just come to light in the recent decade. Various soft robotics designs allow robots to integrate into the natural environment and safely facilitate our daily lives.

## 1.2. Motivation

With all the potentials and advantages elaborated in the previous section, the research and development of the robotic fish with soft actuation structure have just started in the recent decade and is still in the primary stage. Most of the existing laboratory-level soft actuator-based robot (SARs) fish designs could only operate at a certain depth instead of moving along the 3D trajectory. Meanwhile, there are currently no commercial level products.

For the applications of scientific aquatic creature habitats and behaviours observation, an agile robot capable of swimming along a 3D trajectory and closely interact with aquatic creatures with long operating time is necessary. A biomimetic soft robotic fish is capable of achieving such functionality better than servo motor actuated robotic fish and propeller-powered ROVs.

The motivation of the project is to investigate the possible design and performance of soft-actuator-powered robotic fish for the next-generation underwater autonomous robot for aquatic creature study. The significance of the research does not solely come from its potential to provide a new form of interaction with underwater creatures, it could inspire future robotic designs that combine biological structure and robotics to achieve extraordinary functionality. It could also be helpful for the research of hydrodynamics of fish with artificial lateral lines for underwater navigation research and swarm robotics behaviour research. [3]

## 1.3. Aim and Objectives

This project aims to build a robotic fish capable of swimming along the 3D trajectory and fully replacing the servo motor actuators on the traditional biomimetic fish robot with soft actuation structures. Compared with existing similar designs, the project aims to resolve low turning control precision and limitation of operating time due to gas cartridge volume. Hopefully, the research can increase the dexterity of the robotic fish while eliminating the restrictions on depth control and operation time.

Since the project is highly practical and experimental, in-lab manufacturing, experiments, and testing are crucial for delivering the satisfactory outcome of the project. As the UK was in lockdown throughout the period of the project, although the biggest efforts have been made to test the performance of the robotic fish physically, the manufacturing of robotic fish is impossible before the presentation of the project due to the restriction of access to the school facilities and government policy. After discussion with the project supervisor, it was agreed that the physical manufacturing and testing were replaced by virtual simulations on the subsystems, and both the final report and presentation will base on the design and simulation result.

The physical soft robotic fish for real-world testing has been replaced by CAD modelling and necessary hyper-elastic Finite Element Method simulation of the performance of the soft actuator in COMSOL; the testing on the program and performance of the controlling circuit board was replaced by the Proteus simulation. The simulations should provide solid verification of the design for future physical manufacturing.

### 1.4. Literature Review

#### 1.4.1. Existing Robotics Fish Designs

Mustafa Ay et al. [4] developed a robotic fish that actuated by a two-link tail mechanism. Each link is driven by a servo motor. The elevator up-down motion is achieved by using a servo motor to slide the battery along the medial axis with a lead screw mechanism. In the physical testing, the robotic fish can reach a circular turning motion of 0.2792 m. For the two pitch angles choose to test the elevator up-down motion, the robotic fish could achieve effective depth control with the lead screw Centre of Gravity (CoG) adjusting mechanism. Furthermore, the author mentioned that all parts need to be covered with epoxy resin to prevent potential water leakage. The difficulty of waterproofing goes up for the multi-joint actuation structure.

Andrew D. Marchese et al. [5] have developed an Autonomous soft robotic fish capable of escape manoeuvres using fluidic elastomer actuators, a CO2 cartridge, and flow control valves are used to enable escape manoeuvres by releasing a large volume of $CO_2$ gas ranging from a baseline flow of 5 L/m to a maximum flow of 50 L/m into the pneumatic chamber. However, there are several drawbacks to Andrew's design. Firstly, the swimming time is restricted by the gas volume of the CO2 cartridge since the pressurised gas is released into the water after inflating the pneumatic actuator. Secondly, the design of CO2 cartridge and flow control valves could only pressurise one side of the pneumatic actuator instead of pressurising one side while vacuuming another, which may decrease the efficiency of the swimming. Furthermore, most importantly, the design does not provide control on pitch angle for 3D trajectory swimming.

Robert K. Katzschmann et al. [6] have developed a robotic fish driven by a soft hydraulic actuator, the Buoyancy Control Unit, and Dive Planes to enable the fish to dive at the depths of 0-18 m. The fish can achieve an average swimming speed of 21.7 cm/s. A custom-designed unidirectional acoustic communication modem accomplishes the real-time control of the robot, and the distance of successful signal transmission is around 21 m from the robotics fish. Hence, a human diver needs to follow the fish. There is also a fisheye camera placed at the front to record real-time video from the view of fish. The tuning control was achieved by the dive planes at a limited depth. Manual adjustment on the weight by the diver is needed when the fish goes out of the allowed range in order to return to the original depth.

Tiefeng Li et al. [7] developed a fast-moving biomimetic menta ray robotic fish with dielectric elastomers soft actuators. The design took advantage of the surrounding open water as electric ground and boost the battery voltage to 10kV to drive the dielectric elastomers soft actuator. The fish could achieve 0.024 W input power and 10.25% power efficiency, which is comparable to the actual fish with similar size.

Due to the limitation on the maximum page count, the table of comparison of the designs is attached in Appendix 6.1 to provide a more structured view of the pro and cons of existing designs.

### 1.4.2. Soft Actuation Manufacturing and Control

A variety of pneumatic actuation structures, manufacturing and control methods were developed to fit into different needs for actuation. The FDM 3D printing manufacturing method for dual-channel bellows-type actuators with a bottom-up approach was developed by Hong Kai Yap et al. [8]; The actuator can be directly printed with NinjaFlex filament with Shore hardness of 85A, comparing with other designs, the actuator can be wholly printed with FDM 3D printer, which is more accessible than expensive SLA 3D printers. The manufacturing process is simplified comparing with mould casting as well. Although the actuator can withstand a maximum of 400 kPa pressure before failure, the actuation pressure needed is higher than silicon actuators with similar size since the silicone rubbers generally have the Shore hardness below 45A, compared to 85A filament. The shape bellows-type actuators are non-ideal to be placed on robotic fish in terms of hydrodynamic efficiency since the bumpy faces will generate higher drag than streamline body.

Raphael Deimel et al. [9] developed a silicone-rubber-made soft pneumatic actuator called "PneuFlex" that can be manufactured with a two-part mould and reinforcement helix; the manufacturing process is significantly simplified by the novel actuating structure. Except for the FDM 3D printing, Direct in writing (DIW), Selective sintering (SLS), Inkjet printing, and stereolithography technology could be used to manufacture various soft robotic systems materials with different properties.

In terms of the sensing and control of robotic fish actuation structures, Yu-Hsiang Lin et al. [10] embedded a soft sensing mechanism with liquid Gallium-Indium alloy metal in flexible microchannels (eGain) in the soft fishtailing actuators, the morphing of the microchannels will cause the resistance change in the liquid metal, and the strain can be calculated based on the resistance change. Such a flexible sensing mechanism enables closed-loop control robotic swimming based on the sensor measurements. Comparing with other open-loop controlled soft actuators, the precision of the swimming trajectory can be significantly improved.

## 1.5. Scope

The following parts are included in the rest of the report:

In the second section, detailed mechanical designs, and the manufacturing method of some of the key mechanical components will be first introduced. The components include the soft actuators, gear pump, waterproof shell, pectoral fins, and balance control unit. The electronic components design, including water sensor, microcontroller, application board, IMU, and battery pack, are introduced after the mechanical designs.

The implementation and results of three subsystem simulations are included in the third section. The simulation on the application board functionality and simplified control algorithm is elaborated in the first part. And then, the method to perform hyperelastic FEM simulation on the soft actuator performance and simulation results are introduced. At last, the CFD simulation on the hydrodynamic efficiency of the robotic fish design is covered.

Finally, the summary of execution is addressed in the fourth section of the report. The highlights and flows of the design and simulations are analysed in this part. Based on the analysis, future work that can be done to further improve the design and the plans to physically manufacture the robotic fish is further discussed at the end of the report.

## 2. Robotic Fish Design

The assembly of the robotic fish is shown in Figure 2.1 below. The soft robotic fish includes ten main components: Soft actuator, gear pump, waterproof shell, balance control unit, artificial pectoral fins, inertial measurement unit (IMU), battery pack, application board, water sensor, and microcontroller. The design and manufacturing of each component will be elaborated and discussed in this section.

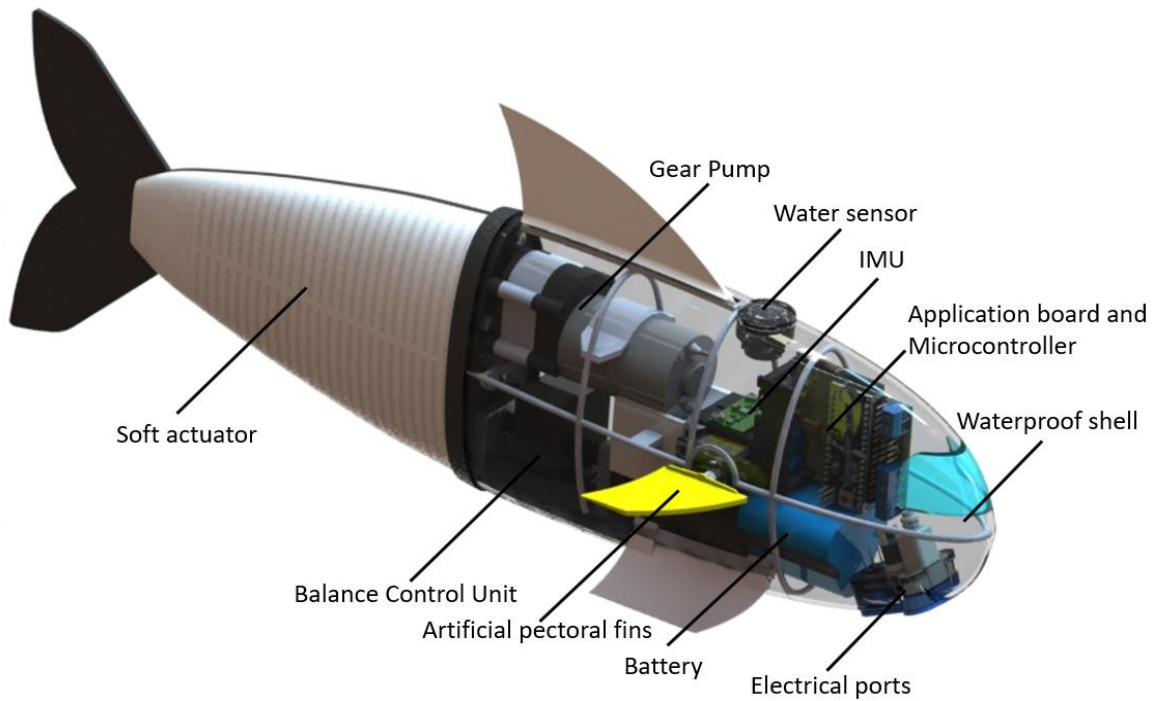

Figure 2.1. The transparent view and components of the robotic fish

## 2.1. Fishtailing Actuators

The soft actuator is the most crucial component in the soft robotic fish design. A deep dive into the available designs was conducted. There are three main design methods: hydraulic, pneumatic, and dielectric elastomer actuated.

The soft pneumatic actuator is ideal for fast turning, but it is extremely difficult to swim along a 3D trajectory since the buoyancy distribution will be affected when the pressurised gas is released to the actuator; the gas volume in the cartridge will limit the operation time, and the replacement of the cartridge is problematic. Dielectric elastomer actuator is a novel design that generates extremely low noise during the operation, although a boost converter is needed to boost the battery voltage to more than 10 kV in and the bending of the actuator is not symmetrical. Thus, it is not suitable for achieving fishtailing behaviour.

The hydraulic actuator was chosen to be the tailing actuator of the fish. Since the liquid could be cycled between the left and right actuator and will not affect the buoyancy distribution, the limitation of the gas cartridge could be eliminated, and the gear pump is relatively simple to operate. Figure 2.2 shows the assembly of the soft actuator; the actuator consists of 4 parts: (1) left soft fluid actuators, (2) right soft fluid actuators, (3) plastic mount, (4) constrain layer with curvature sensor.

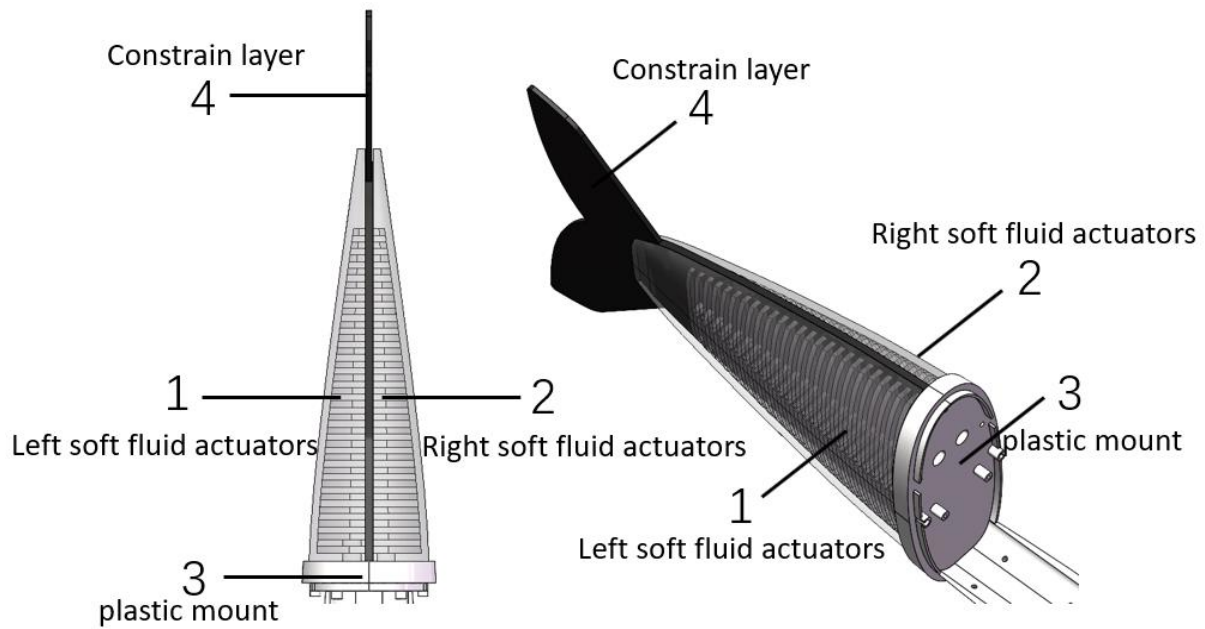

Figure 2.2. Assembly of the soft fishtailing actuator

The soft actuator functions by pressurising and depressurising the soft actuators with liquid by a gear pump, the fluidic channels within the actuator will be expanded and compressed, and the constrain layer will be bent as it is inextensible in length. By cycling the fluid between left and right actuators, they are pressurised and depressurised separately; the constrain layer will be bent in both directions. Through such a mechanism, the fishtailing behaviour can be imitated. [6]

Ecoflex 30 Silicon rubber will be used to fabricate the soft fluid actuator by mould casting. The mould design is shown in Figure 2.3; the design consists of 3 separate parts: top mould, bottom, and plug for generating the water in/outlet hole. All three parts will be manufacturing by a Stereolithography 3D printer so that the top and bottom can be precisely assembled to each other. Because removing the support from the 3D printed part will significantly affect the casting part's precision and smoothness, the parts' design has avoided the occurrence of undercut geometry so there will not be any support generated during the 3D printing. Liquid silicon will be injected from the top of the mould, and after which the mould will be placed in a vacuum container to remove the bubbles; if the liquid silicon level goes down, repeat the previous step several times until the level no longer drops. After curing in the oven, the silicon-made soft actuator can be fabricated.

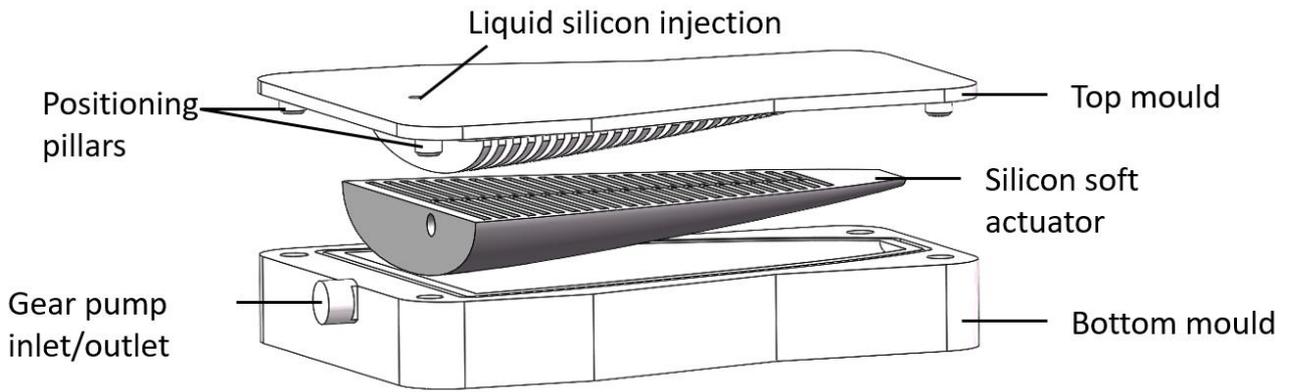

Figure 2.3. Mould casting design for fabricating the soft actuator

2 Spectra Symbol Flex Sensors [11] are placed in the middle of the constraining layer to provide control feedback by measuring the bending angle of the actuator. The sensors will produce 10k Ohm resistance when the constraining layer is flat, and the resistance will increase to up to 110k Ohms as the bending angle increases. Each sensor could only provide reliable measurement in 1 direction, so two sensors will be placed reversely to measure the bending angle in both directions.

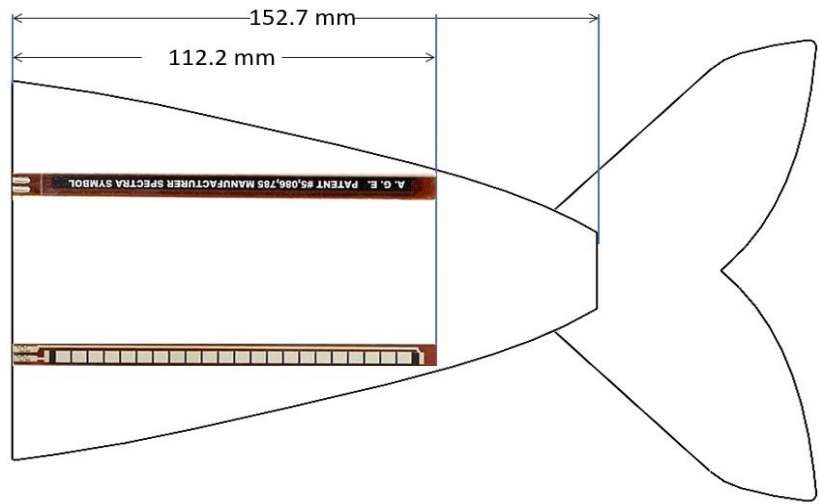

Figure 2.4. Sensor position on middle constrain layer of the actuator

The sensor manufacturer does not provide an accurate resistance-curvature relationship in the datasheet. The sensors will have to be tested physically to validate the accuracy and range of measurement.

## 2.2.    Gear Pump

The decision was made to use a gear pump to cycle the liquid between left and right the soft actuator. The crucial reason for choosing a gear pump is output is directly proportional to the motor speed, which means the gear pump could deliver smooth, pulse-free, and invertible flow comparing with piston pumps and peristaltic pumps. The rate of flow and pressure generated by gear pumps is generally higher comparing with centrifugal pumps and diaphragm pumps with similar power ratings. The compact volume could also fit into the body of the fish.

After conducting research on available gear pumps online, DHE 385 micro gear pump [12] is found to be the optimal option to drive the fishtailing actuator; the gear pump uses an SRC-385MP-2165-54DV motor with 12 V rated voltage and 0.8 A maximum rated load current, it can produce 2L/minute liquid flow and achieve maximum 2.45 bar pressure. However, the water inlet and outlet of the pump are placed on the side of the gear pump (shown in Figure 2.5), which is non-idea for the connection with fishtailing actuator.

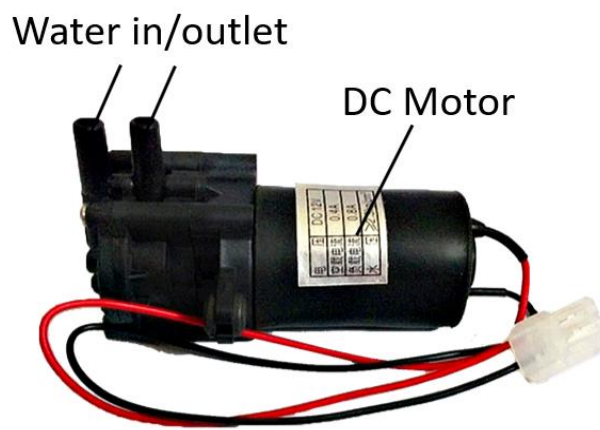

Figure 2.5. DHE 385 Micro gear pump [12]

Reverse engineering was done on the DHE 385 Micro gear pump by pulling it apart and investigating the working principle. The new design based on DHE 385 moved the liquid inlet and outlet from the side of the gear pump to the top, which enables a direct connection between fishtailing actuator and gear pump. (Figure 2.6)

There are two sets of gears within the gear pumps. The spur gear set that is directly connected to the motor has a gear ratio of 0.25 between input and output shaft, which could increase the rated torque of the motor by four times. Bevel gears are used as gears for pumping liquid, which means the gears are spiral along the axial direction of the gear. The design enables the gear tooth to join and disengage gradually with each other throughout the rotation. Thus, the noise and vibration during the operation could be reduced comparing with spur gears.

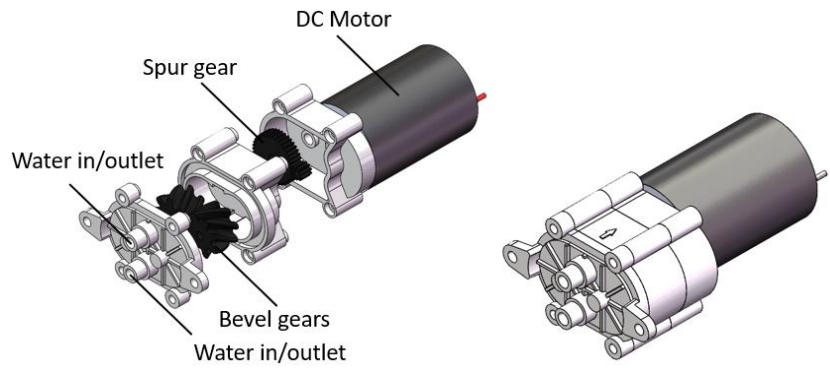

Figure 2.6. Redesigned gear pump and its explosive view

### 2.3. Waterproof Shell

The shell protects all the electronic components from contact with water during the operation. It consists of 5 main parts: pelvic fins, dorsal fin, rigid supporting frame, shell, and electrical ports.

The main functions of the pelvic fins and dorsal fin are to stabilise the fish during swimming and prevent the fish from rolling during the turning. The electronic components are assembled on the 3D printed rigid frame as shown in Figure 2.7, except for the water sensor and electrical ports are assembled on the semi-soft shell, which is printed with 90A Shore hardness resin. The shell is supported by the rigid supporting frame. During the assembly, after all the electronic components are assembled and constrained, the supporting structure and shell can be easily assembled to each other, after which applying waterproof glue, the shell is properly sealed. The transparent window at the front of the shell enables the fisheye camera to be placed on the fish.

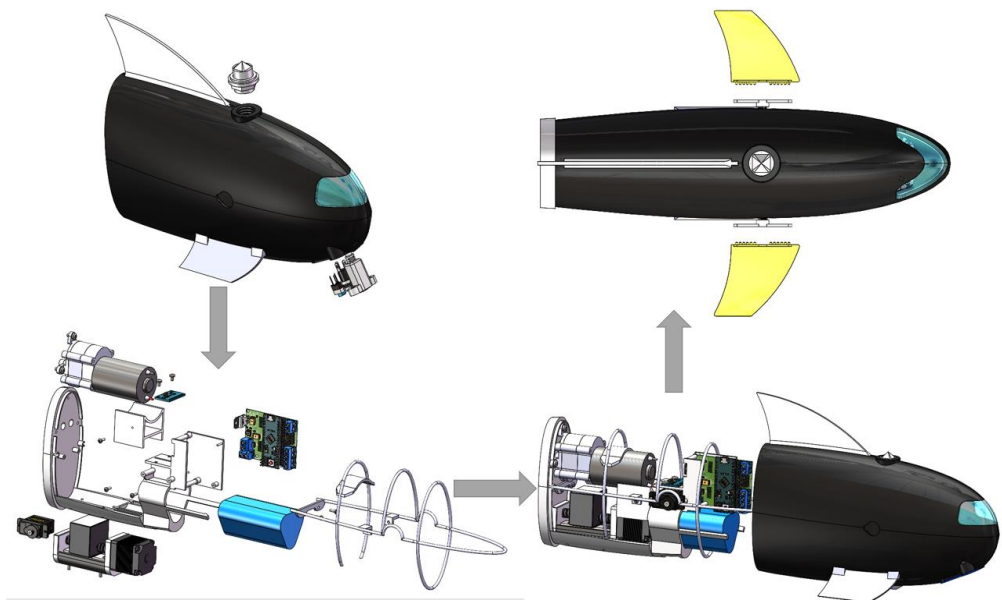

Figure 2.7. Assembly flow of the shell

The electrical port consists of a Micro USB connector, a 5.5 DC power jack, and the main power switch. The Micro USB port is connected to the USB port on the Arduino microcontroller with an extension cable for uploading the program to the microcontroller and acquiring sensor data with the serial monitor. The DC power jack is connected with the battery for charging, and the main power switch is also connected between the battery and the application board. The electrical ports are covered by a removable rubber cap to prevent the water from entering the shell through the gap on the ports.

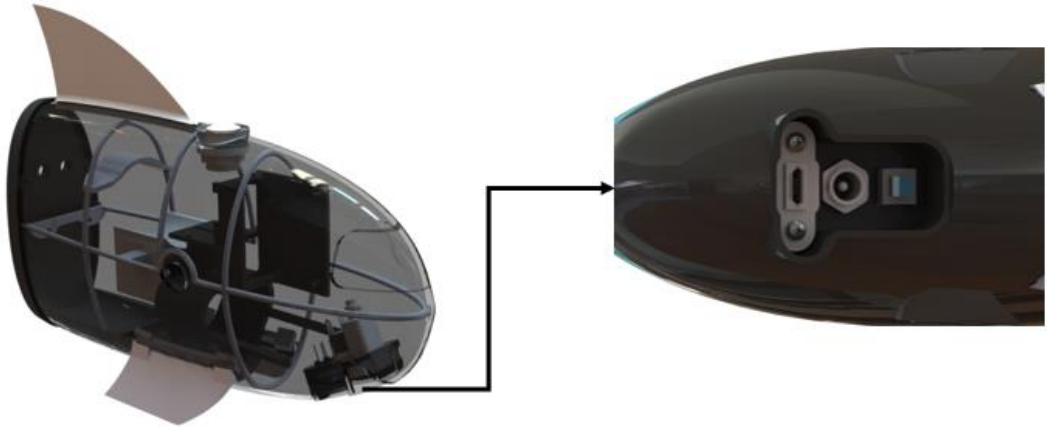

Figure 2.8. Waterproof shell design and electrical ports on the shell

## 2.4. Artificial Pectoral Fins and Balance Control Unit

In order to allow the robotic fish to swim along the three-dimensional trajectory, artificial pectoral fins and a balance control unit (Figure 2.9) are used to adjust the pitch angle of the robotic fish.

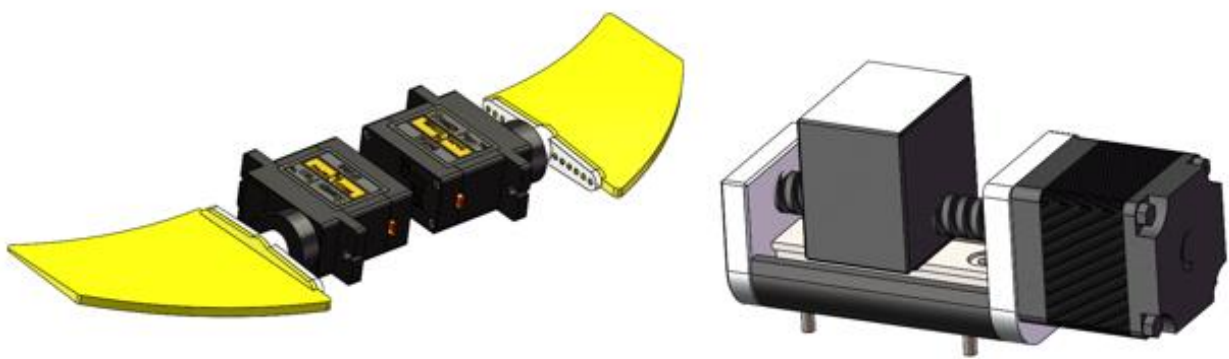

Figure 2.9. Artificial pectoral fins (Left) and Balance control unit

Two Kingmax KM1850MD [13] waterproof servo motors drive the artificial pectoral fins, each of the motors weighs 18 gram and capable of delivering a maximum of 2.8 $kg \cdot m$ torque at 5 V, the operating angle of each motor is 0 to 90 degree at the pulse width of $1000 - 2000 \; \mu sec$, all these features allow the servo motors to be controlled by Arduino microcontroller directly and operated at high hydro mechanical resistance underwater environment. The servo motors can respond rapidly to the change in the turbulence and provide dynamic lift force to the swimming like aeroplanes' wings. The turning response can be faster by rotating left and right pectoral fins in different directions than adjusting the amplitude of the left and right fishtailing motion.

The balance control unit consists of a metal driven by a lead screw and a two-phase hybrid micro stepper motor since stepper motors could provide simple and precise position tracking, the step angle of the stepper motor can be calculated as:

$$step\ angle = \frac{360}{2 \times Phase \times pole} = \frac{360}{4 \times 50} = 1.8° \qquad (2.1)$$

The relationship between the rotational speed of the stepper motor and the linear speed of the metal can be driven as:

$$v = N_L \times 0.002 \qquad (2.2)$$

where 0.002 is the lead of the screw and N is the rotational speed of the screw, and v is the linear speed of the metal.

It controls the pitch angle by sliding the metal along the lead screw, resulting in horizontal adjustment on the centre-of-gravity of the robotic fish.

## 2.5. Water Sensing

FR-IR12 reflective water sensing module [14] is used to detect if the robotic fish is underwater or has reached the water's surface. It is placed on the top of the robotic fish nearby the dorsal fin.

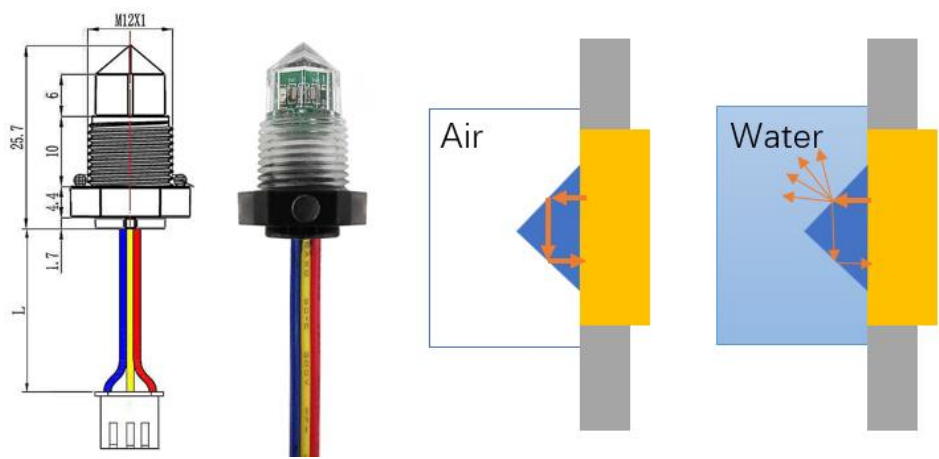

Figure 2.10. FR-IR12 water sensor [14] and sensor function diagram

Comparing with traditional mechanical water sensors, FR-IR12 detects the water through light reflection. The sensor has a prism-shaped probe to reflect the light from an internal luminous diode to the receiver. If the probe is placed in the air, the prism will reflect most of the light to the receiver and outputs a logical high. When the probe is placed in the water, most of the light will scatter in the water; in this case, the receiver will pick up significantly fewer light signals, and the sensor will output logical low.

The compact and reliable sensor design is an excellent choice to be placed on the robotic fish not only because of the space problem, the waterproofing of the robotic fish will also be significantly simplified comparing with traditional mechanical water sensors.

## 2.6. Microcontroller and Application Board

The available space inside the robotic fish is extremely limited. Ensuring that the controller and motor drive board are as compact as possible while the IOs are sufficient for the system is crucial in the design.

Arduino Micro is chosen to be the controller of the robotic fish. It is designed based on Atmel ATmega32U4 8-bit Microcontroller, which containing 20 IO pins (including 7 PWM outputs and 12 analogue inputs); serial pins onboard allow the usage of the IMU. The size of the board is only 48 mm by 18 mm, which is ideal to be placed inside the fish. [15] The Arduino is mounted on the application board through the onboard headers, the size of the application board is 45 mm by 45 mm, components are placed on both sides of the board, the top and bottom layer PCB design is shown in Figure 2.12, and the 3D view is shown in Figure 2.11. Fully schematic diagram that shows the detailed wiring is attached in the Appendix 6.2.

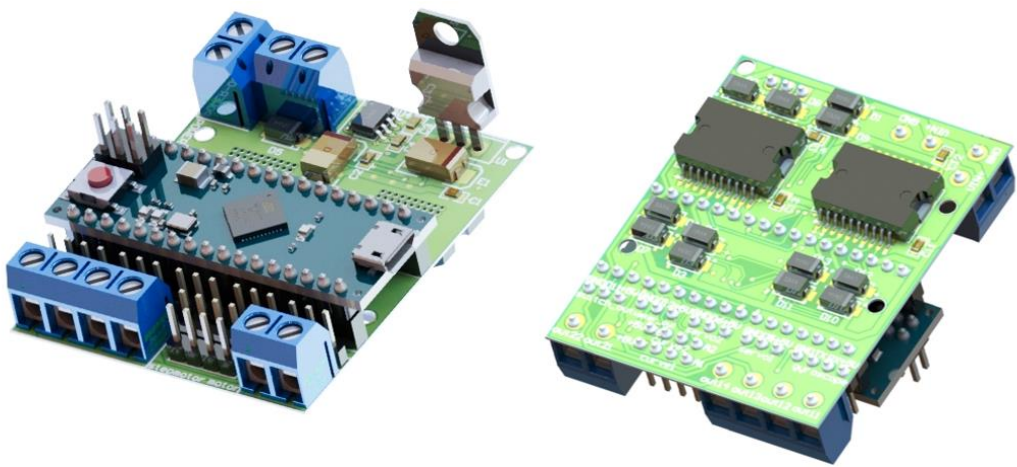

Figure 2.11. 3D view of the application board

The application board mainly consists of 4 parts: 2 H-bridge modules with optical isolation, 12V to 5V step-down voltage converter, current monitor circuit, and connection ports and headers. The wiring diagram of the application board is also shown in the Appendix 6.2.

Polygon Pour connection is used to reduce the mutual capacitance and inductance to improve the signal integrity. Headers for the motors and sensors are placed at the bottom of the board. Most of the heat will be generated at L298P dual H-bridge modules [16] and the L7805 power regulator [17]. They are placed at the top so that the wiring will not affect heatsink assembly and performance. Since the thermal conductivity of water is higher than air, thermal management should not be problematic.

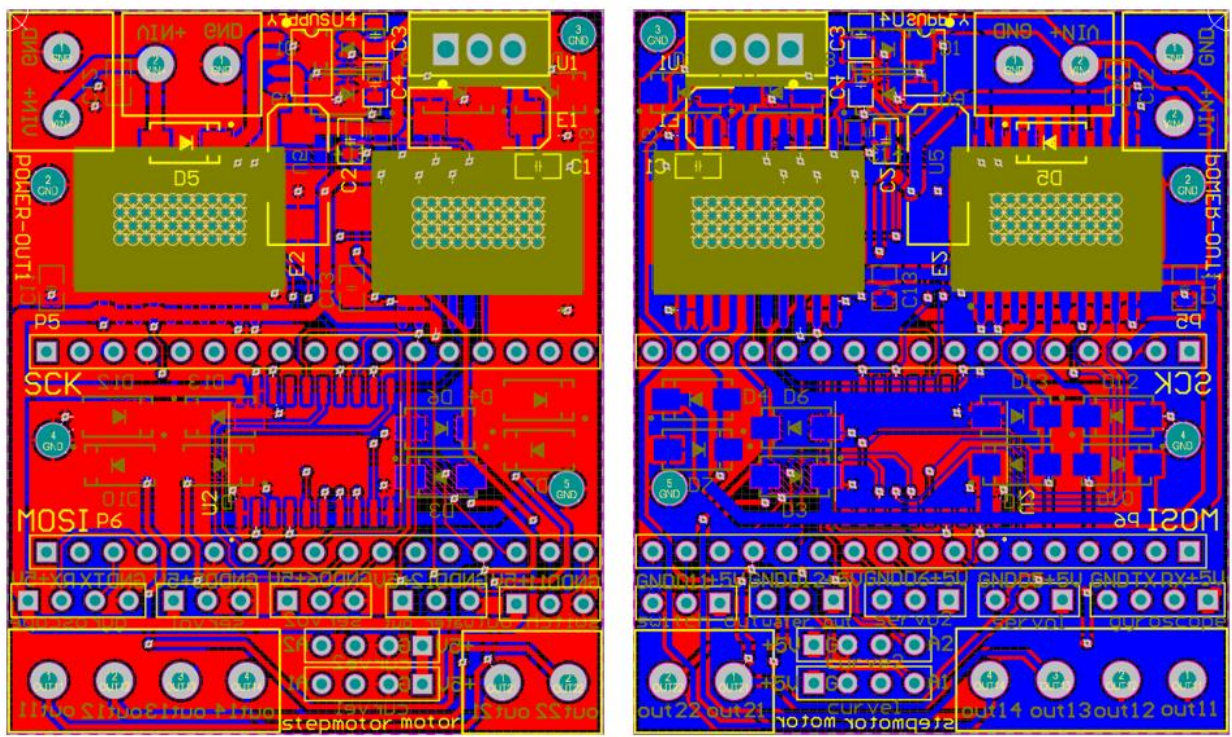

Figure 2.12. Top and bottom layer of application board

2 L298P H-bridge ICs are used to drive the DC and stepper motors; a SN74AHC244DW optical coupler IC [18] provides isolation between Arduino outputs and the H-bridge circuit, the schematic diagram of the L298P modules is shown below.

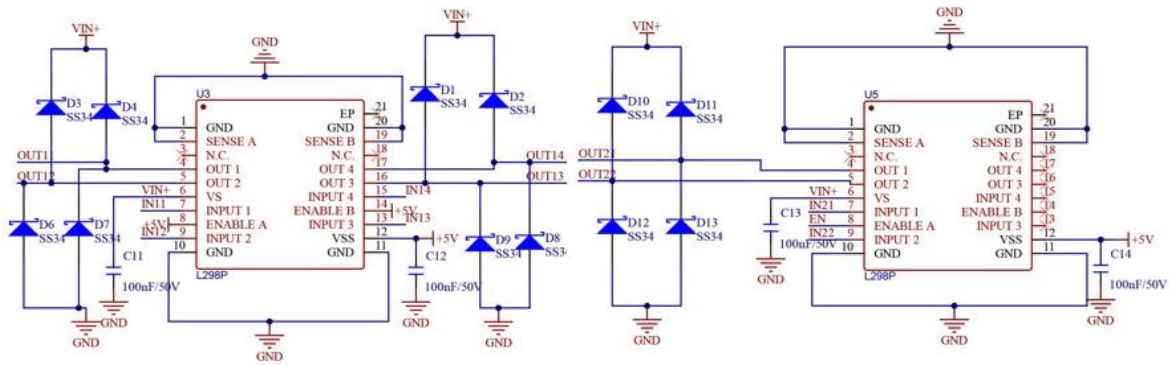

Figure 2.13. Schematic diagram of the L298P modules

The input voltage from the battery is 12V, to supply 5V to the Arduino Microcontroller and servo motors, an L7805 positive voltage regulator IC is used to provide a constant 5V power supply to Arduino and servo motors. The IC can deliver a maximum of 1.5A current with a proper heatsink [17], which is more than the total current rating of the Arduino and servo motors.

An ASC712 Hall Effect-based linear current sensor [19] is used to monitor the real-time current outputs from the battery. The relationship between sensed current and output analogue voltage is shown in Figure 2.15. The numerical relationship at 25 degree Celsius can be derived approximately as:

$$V_{OUT} = 0.19 \times I_P + 2.55 \tag{2.3}$$

The build-in 10-bit Analogue-Digital Converter will divide the 0-5V analogue input voltage into 1024 steps. Thus the numerical relationship at 25 degree Celsius between the digital input and sensing current can be further derived as:

$$Digital\ Out = 40.92 \times I_P + 511.5 \tag{2.4}$$

The schematic diagram of the voltage regulator and current sensing circuit is shown in Figure 2.14.

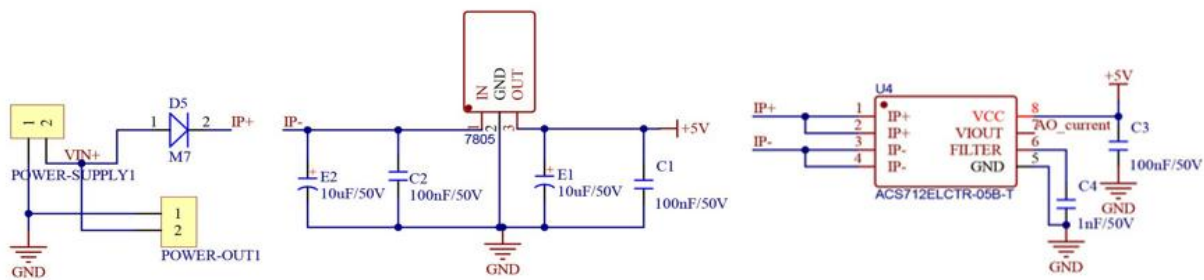

Figure 2.14. Schematic diagram of the voltage regulator and current sensing circuit

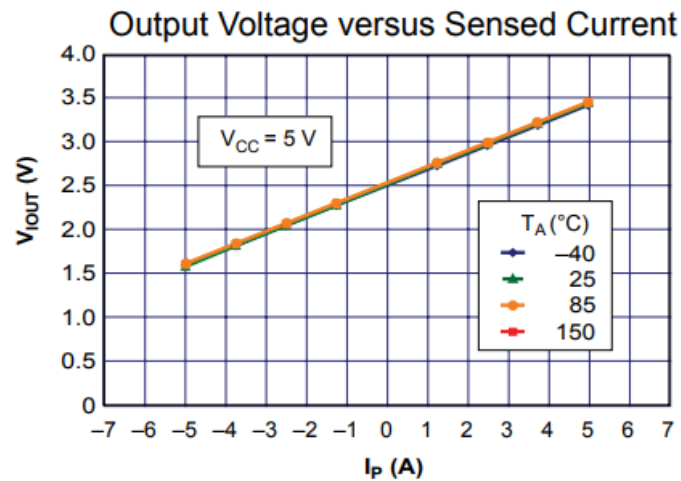

Figure 2.15. Output voltage versus sensed current relationship of ASC712 current sensor [19]

## 2.7. Position Tracking Module

Since the robotic fish is designed for swimming along a 3D trajectory, an Inertial Measurement Unit (IMU) is necessary for tracking the real-time position of the robotic fish. MPU 6050 [20] integrates a 3-axis accelerometer and a 3-axis gyroscope on a single IC, and it is also called 6 degrees of freedom motion tracking device. MPU 6050 also has a temperature sensor and a Digital Motion Processor (DMP) that correlates the data from the accelerometer and gyroscope.

MPU 6050 could provide communication with Arduino through I2C protocol by connecting SCL and SDA pins onboard to the corresponding SCL and SDA pins on Arduino Micro, the Arduino IDE build-in wire library can directly read and write data through I2C bus.

For the accelerometer, there are four programmable scale range: ±2g, ±4g, ±8g, and ±16g. Each scale represents the maximum possible reading of acceleration from the accelerometer. The acceleration in x, y, and z direction can be calculated by:

$$a_x = s \times \frac{d_x}{32768} \qquad (2.5)$$

$$a_y = s \times \frac{d_y}{32768} \qquad (2.6)$$

$$a_z = s \times \frac{d_z}{32768} \qquad (2.7)$$

where vector a is the numerical acceleration in the direction, s is the chosen scale range, d is the axis acceleration data acquired from MPU 6050 module, and 32768 is the maximum possible reading from the 16-bit ADC module onboard.

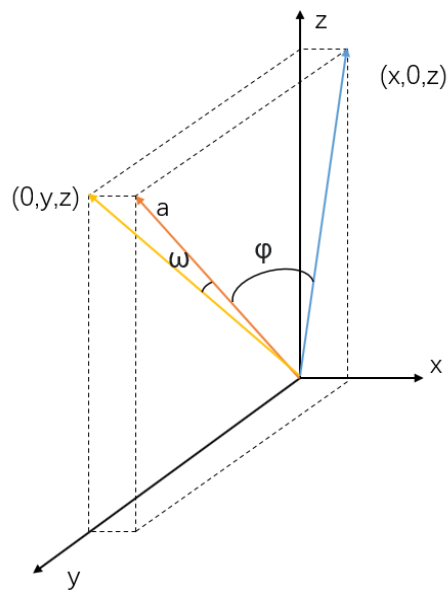

Figure 2.16. Roll and Pitch angle

The roll and pitch angle can be calculated through the acceleration. A vector represents the overall acceleration of the robotic fish, assuming the robotic fish is swimming at a constant speed, the vector should point in the positive direction of the z-axis. If the robotic fish rotates, the acceleration will not point in the positive direction of the z-axis, as shown in Figure 2.16, the roll angle ɸ (blue) is the angle between vector *a* and its projection on the plane (x, 0, z), the pitch angle ω (yellow) is the angle between vector *a* and its projection on the plane (0, y, z), the angle can be calculated with trigonometrical relationships by:

$$\varphi = \cos^{-1}\left(\frac{\sqrt{x^2 + z^2}}{\sqrt{x^2 + y^2 + z^2}}\right) \tag{2.8}$$

$$\omega = \cos^{-1}\left(\frac{\sqrt{y^2 + z^2}}{\sqrt{x^2 + y^2 + z^2}}\right) \tag{2.9}$$

Since the arccosine function can only return a positive number, the roll angle needs to be negative if y is positive, and the pitch angle needs to be negative if x is negative.

It is difficult to calculate absolute yaw angle without the reference of earth magnetic field measurement, but the change in yaw angle, which is the angular velocity can be easily acquired, and the measurement of angular velocity is sufficient for the application.

For the gyroscope, there are also four programmable scale range: 250 degrees/s, 500 degrees/s,

1000 degrees/s, 2000 degrees/s. From the same principle, the angular velocity in x, y, and z direction can be calculated by:

$$\omega_x = s \times \frac{d_x}{32768} \quad (2.10)$$

$$\omega_y = s \times \frac{d_y}{32768} \quad (2.11)$$

$$\omega_z = s \times \frac{d_z}{32768} \quad (2.12)$$

Where $\omega$ is the numerical angular velocity in the direction, s is the chosen scale range, d is the axis angular velocity data acquired from MPU 6050 module, and 32768 is the maximum possible reading from the 16-bit ADC module onboard.

A complementary filter can be implemented to combine the roll and pitch angle calculated by the accelerometer measurements and direct measurement from the gyroscope. Gyroscope could normally provide accrete measurement since it is not affected by linear motion, and thus it normally has a higher weight when merged. Still, the angle calculated by the accelerometer can offset the error caused by drifting in the gyroscope from a long-term perspective.

Although there is a DMP on MPU 6050 module, the output signal is still highly likely to be noisy. Kalman filter is ideal for eliminating the noise in the readings. It can be implemented directly through the Kalman filter library in Arduino IDE.

### 2.8. Battery

The battery pack consists of 3 18650 protected lithium-ion rechargeable cells connected in series, each of the cell has a charging voltage of 4.2 V and the nominal voltage is 3.7 V, 3 of the batteries connected in series provide a maximum voltage of 12.6 V, and the nominal voltage of 11.1 V, depends on the battery level. The 18650 cells normally have a capacity between 1800 mAh to 3500 mAh [21]. In order to maximise the operating time of the fish, Panasonic NCR18650B 3400 mAh battery cell will be used to manufacture the battery pack.

Since the space within the fish body is very limited, three cells are placed in a triangle shape to minimise occupied volume, PVC heat shrink packaging could prevent the battery from short-circuit caused by water leakage, the 3D rendering image and example image provided by the battery pack manufacturer is shown in Figure 2.17 below.

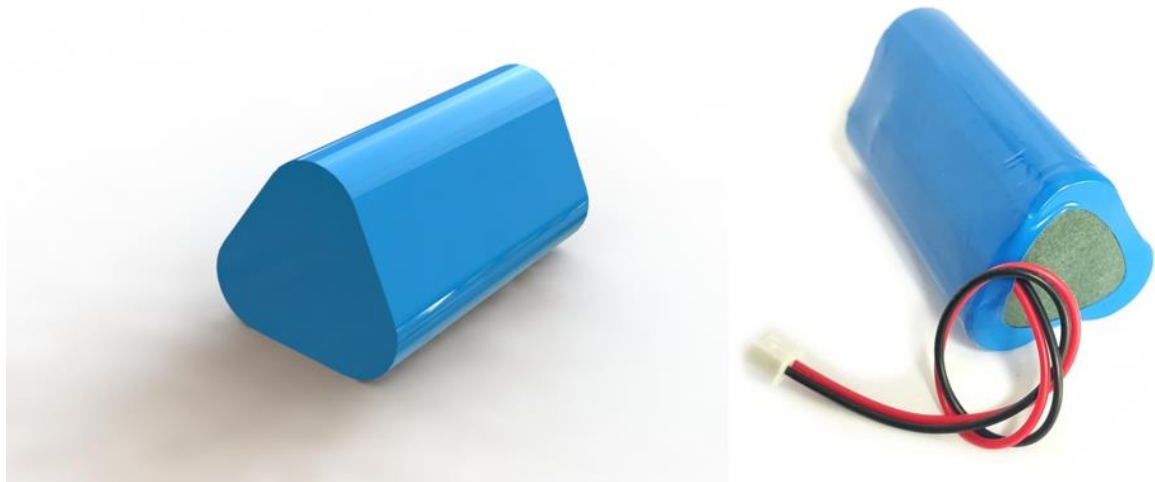

Figure 2.17. 3D rendering image (left) and example image provided by the battery pack manufacturer (right)

Panasonic NCR18650B battery has a built-in protection circuit to prevent the battery from overcharge, over-discharge, and short circuit. Such a feature makes the battery safer to be used in the robotic fish since a short circuit caused by water leakage is highly likely to happen during the testing. The circuit can effectively protect the controller board and other electronic components.

## 3. Subsystem simulation

Because of the outbreak of the coronavirus, at the current stage, physical testing and manufacturing of the subsystem are challenging to achieve; thus, two methods will be used to conduct testing on the subsystem development: Hyper-elastic FEM simulation on the soft actuator based on COMSOL Multiphysics and circuit control simulation based on Proteus Design Suite.

### 3.1. Simulation on Application Board and Simplified Control Algorithm

Proteus Design Suite is chosen to simulate the circuit and program's performance due to its compatibility with the Arduino microcontroller.

The circuit is simplified to the version in Figure 3.1, and it includes 2 L298P ICs, an ATmega32U4 microcontroller, a stepper motor, a DC motor, two servo motors, and five switches as controller input. An additional oscilloscope and probes are also added to the system to observe and verify the control and output signal.

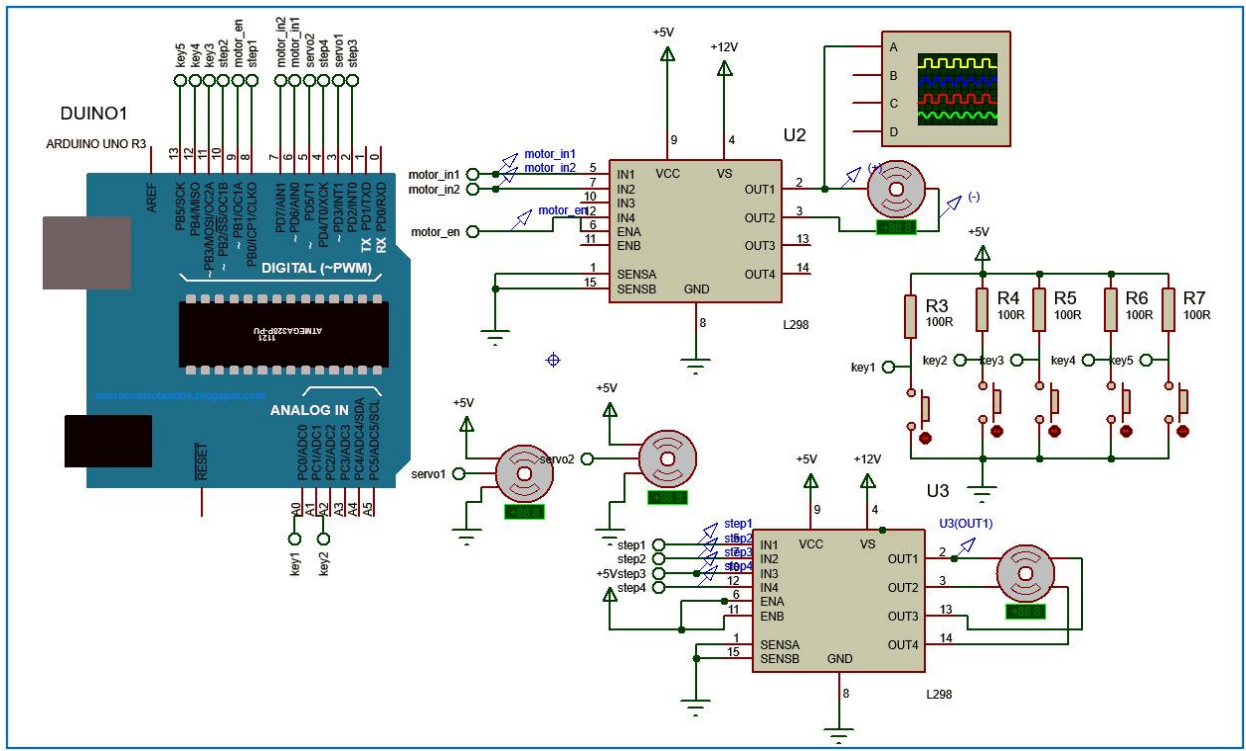

Figure 3.1. Proteus circuit diagram for simulation

The analogue input in the system is difficult to simulate in the simulator. The simplified control of the robotic fish with five push buttons ignores the control feedback from the flexible curvature sensors and the IMU. However, it is sufficient to validate the functionality of the application board design. The implementation schematic of the simplified control algorithm diagram is shown in Table 3.1.

Since the DC motor is used to drive the gear pump to cycle water between left and right soft actuator, and the angular speed is proportional to the input voltage for the DC motors, the motion of the DC motor can be expressed by a function of input voltage and time, for the Straight Motion in Table 3.1, the expression can be derived as:

$$V_{Straight}(t) = V_A \times \sin(\omega t) \tag{3.1}$$

where the $V_A$ is the desired maximum voltage input to motor and $\omega$ is the fishtailing frequency in rad/s. Assuming $V_A$ is the maximum voltage 12 V and the frequency is 1 Hz, Figure 3.2 shows how the straight swimming can be achieved by $V_{Straight}(t)$ function with the aid of FEM simulation results, which will be discussed in section 3.4.

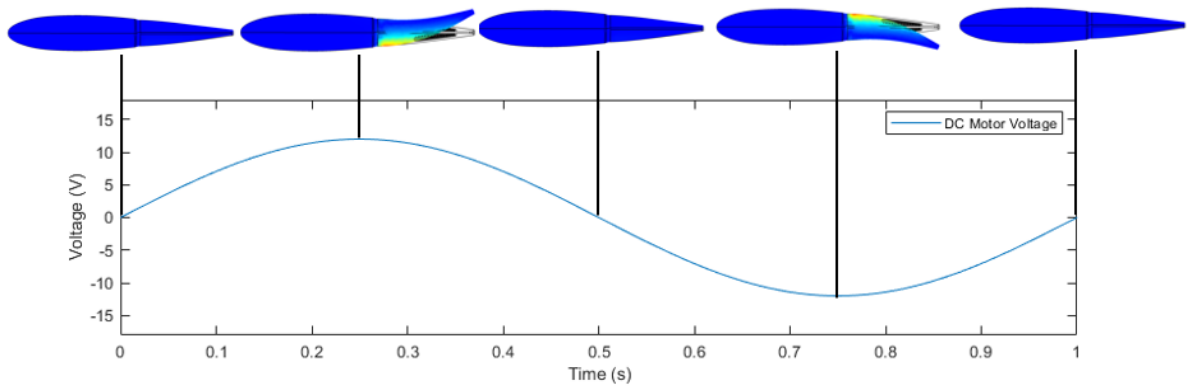

Figure 3.2. DC motor voltage graph for straight swimming fishtailing movement

The left and right turning can be achieved by adjusting the amplitude of forward and reverse rotation, the expression of input voltage for left and right turning can be expressed as:

$$V_{Left}(t) = \begin{cases} V_{A1} \times \sin(\omega t), & 2n > t > 2n+1 \\ V_{A2} \times \sin(\omega t), & 2n+1 > t > 2(n+1) \end{cases}, n = 1,2,3,4,5,\ldots,\infty \quad (3.2)$$

$$V_{Right}(t) = \begin{cases} V_{A1} \times \sin(\omega t), & 2n > t > 2n+1 \\ V_{A2} \times \sin(\omega t), & 2n+1 > t > 2(n+1) \end{cases}, n = 1,2,3,4,5,\ldots,\infty \quad (3.3)$$

where the $V_{A1}$ and $V_{A2}$ is the desired maximum voltage inputs in positive and negative directions to motor and $\omega$ is the fishtailing frequency in rad/s. According to the design, a positive voltage will result in depressurisation in the left actuator and pressurisation in the right actuator and vice versa, and thus for $V_{Left}$, $V_{A1}$ should always larger than $V_{A2}$ in order to turn left. From to the same principle for $V_{Right}$, $V_{A1}$ should always smaller than $V_{A2}$ in order to turn right. Assuming for $V_{Left}(t)$, $V_{A1}$ is 12 V and $V_{A2}$ is 4 V, and for $V_{Right}(t)$, $V_{A1}$ is 4 V and $V_{A2}$ is 12 V. The frequency for both of the function is 1 Hz. The graph shows how the left and right turn can be achieved.

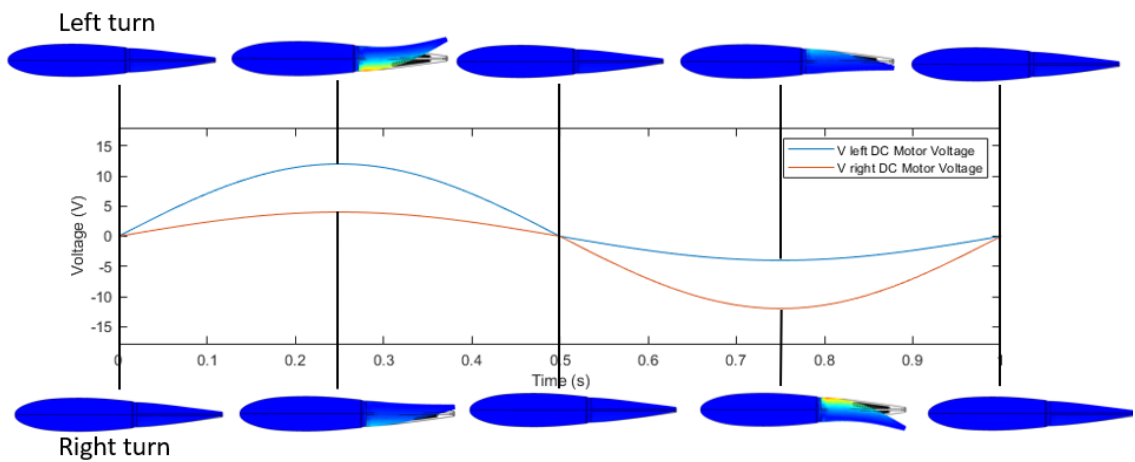

Figure 3.3. DC motor voltage graph for left and right turning fishtailing movement

Table 3.1. Table of simplified motion control

| Control Button | Movement | Motion of Motors |
|---|---|---|
| Key1 | Straight Forward 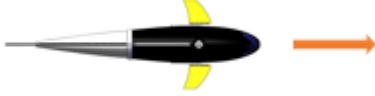 | DC Motor: Motion Straight $V_{Straight}(t)$<br>Stepper Motor: Rotating from current position to 0 degrees position<br>Left Servo Motor: Rotating to 0 degrees position<br>Right Servo Motor: Rotating to 0 degrees position |
| Key2 | Left Turn 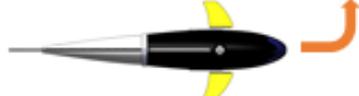 | DC Motor: Motion Left $V_{Left}(t)$<br>Stepper Motor: Stepper Motor: Rotating from current position to 0 degrees position<br>Left Servo Motor: Rotating to +30 degrees position<br>Right Servo Motor: Rotating to -30 degrees position |
| Key3 | Right Turn 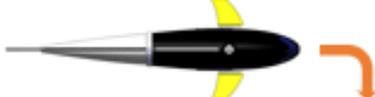 | DC Motor: Motion Right $V_{Right}(t)$<br>Stepper Motor: Rotating from current position to 0 degrees position<br>Left Servo Motor: Rotating to -30 degrees position<br>Right Servo Motor: Rotating to +30 degrees position |
| Key4 | Elevator up 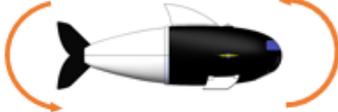 | DC Motor: Motion Straight $V_{Straight}(t)$<br>Stepper Motor: Rotating from current position to -1080 degrees position<br>Left Servo Motor: Rotating to -30 degrees position<br>Right Servo Motor: Rotating to -30 degrees position |
| Key5 | Elevator down 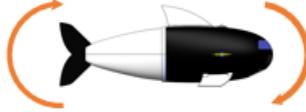 | DC Motor: Motion Straight $V_{Straight}(t)$<br>Stepper Motor: Rotating from current position to +1080 degrees position<br>Left Servo Motor: Rotating to +30 degrees position<br>Right Servo Motor: Rotating to +30 degrees position |

The detailed implementing code in Arduino IDE is attached in the Appendix. The sinusoidal change in DC motor voltage can be achieved by changing the duty ratio of the Pulse Width Modulation (PWM) waveform. Timer library is utilised to generate an interrupt every 1 millisecond to change the duty ratio according to the predefined sinusoidal functions. According to equation (3.2) and (3.3), two sinusoidal functions are defined with the same frequency and different amplitude to generate asymmetrical fishtailing to turn the direction.

The five pushbuttons in the simulation interface can be clicked to control the operating mode to switch between 5 modes in Table 3.1.

Although the half-stepping control method could increase the accuracy of stepper motor control, the full-stepping method is used to implement stepper motor control since for lead screw structure, one revolution will only result in 2 mm linear motion, half-stepping seems redundant in such application.

After compiling the code, importing the hex program file to the microcontroller, and run the simulation, the motor rotation status will be displayed in the simulation. The oscilloscope will also display the input and output signal to further validate the circuit design.

### 3.2.    Proteus Simulation Results

When the robotic is swimming straight, which means $V_{Straight}$ has been applied to the DC motor on the gear pump. The input signal that the microcontroller applied to the L298N module is shown in the Figure 3.4 below. The signal that coloured green is the PWM signal measured by the probe in enable pin on the L298P module, and the signal coloured in blue and red are the signals which control the output voltage polarity. It can be observed that the duty ratio of the input PWM signal changes according to the sinusoidal relationship, and the polarity changes every half cycle. The waveforms meet the design principle.

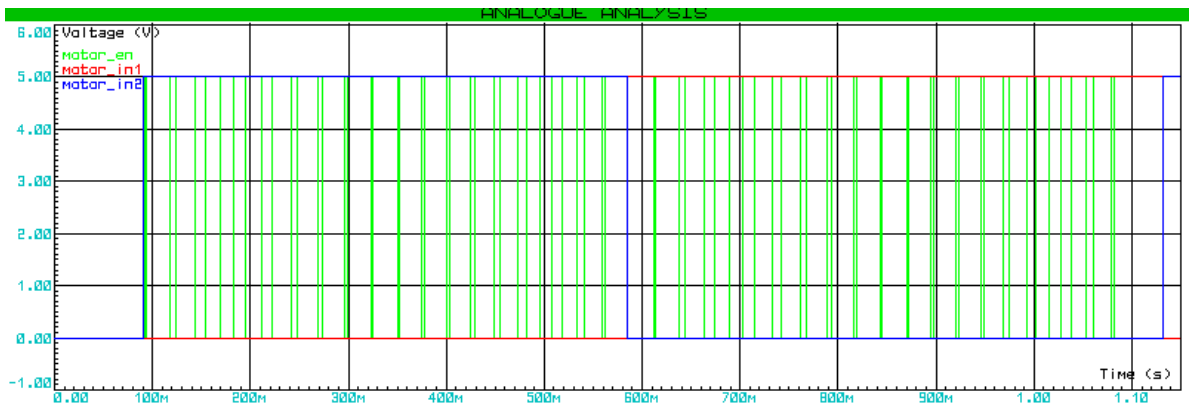

Figure 3.4. $V_{Straight}$ control input signals to L298P

The output voltage applied to the DC motor is shown in the Figure 3.5 below.

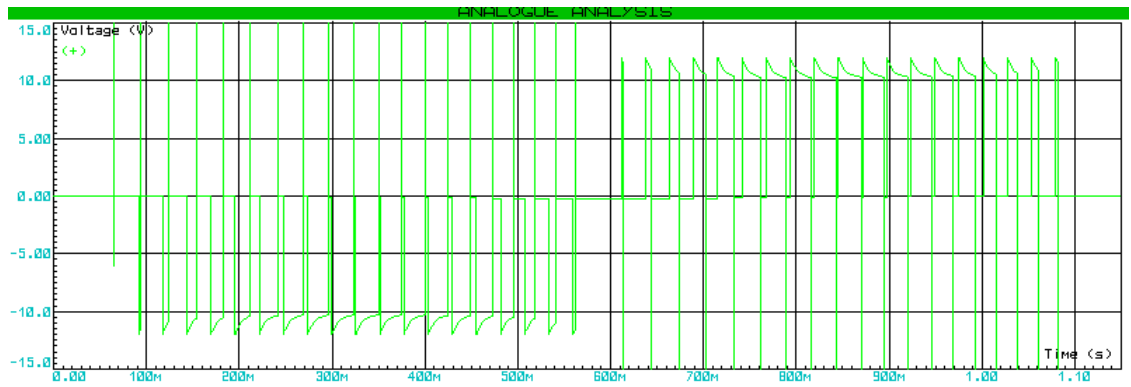

Figure 3.5. L298P output voltage when $V_{Straight}$ is inputted

It can be observed from the output that within half period, the duty ratio gradually increases from zero to the maximum at the centre and then gradually decrease to zero again, which shows the same pattern as the control signal, the voltage is negative at first half cycle and positive at the other half. There is larger overshooting at the turn-off instant of transistors in L298P. Although the voltage of overshooting is large, the time of overshooting is measured to be less than 1 microsecond in the simulation, such overshooting will not have any impact on the system so it can be ignored. To better visualise the signal, the "filloutliers" function in MATLAB can be used to eliminate the overshooting data while keeping the signal integrity. The signal after applying the filter is shown in the Figure 3.6 below.

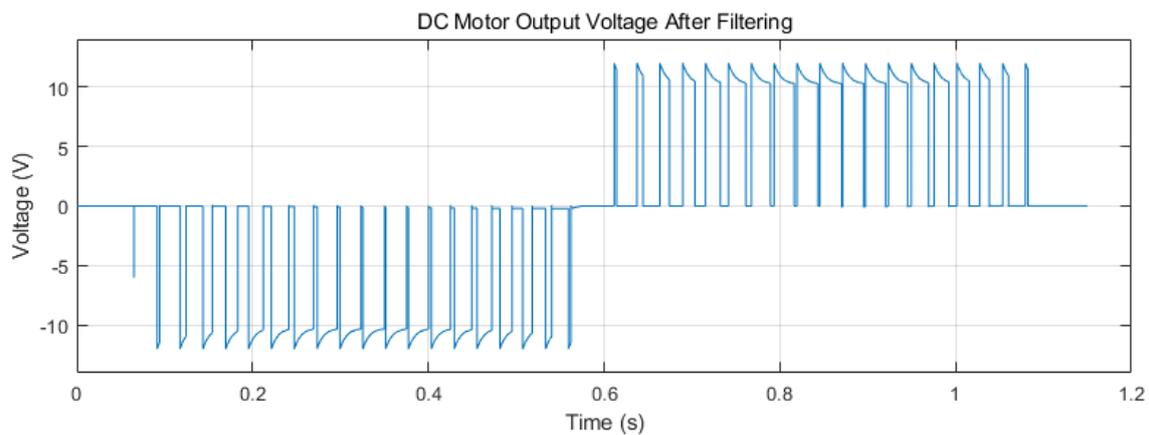

Figure 3.6. L298P output voltage when $V_{Straight}$ is inputted after filtering

Result of simulation for $V_{Left}$ and $V_{Right}$ shows a similar pattern as $V_{Straight}$. Comparing with $V_{Straight}$, the maximum duty ratio for $V_{Left}$ at positive voltage is smaller comparing with duty ratio at a negative voltage, which meets with the requirement described in section 3.1 on the numerical expression of $V_{Left}$ that $V_{A1}$ should larger than $V_{A2}$.

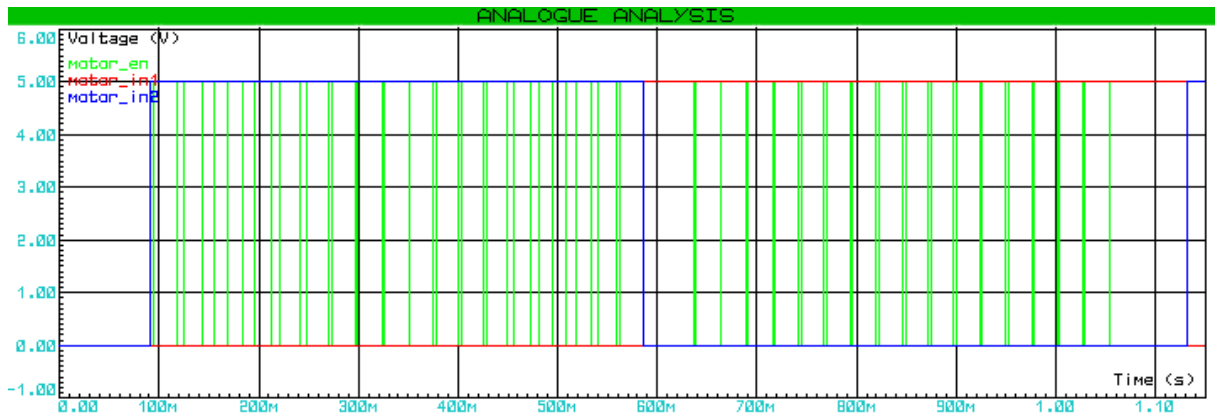

Figure 3.7. $V_{Left}$ control input signals to L298P

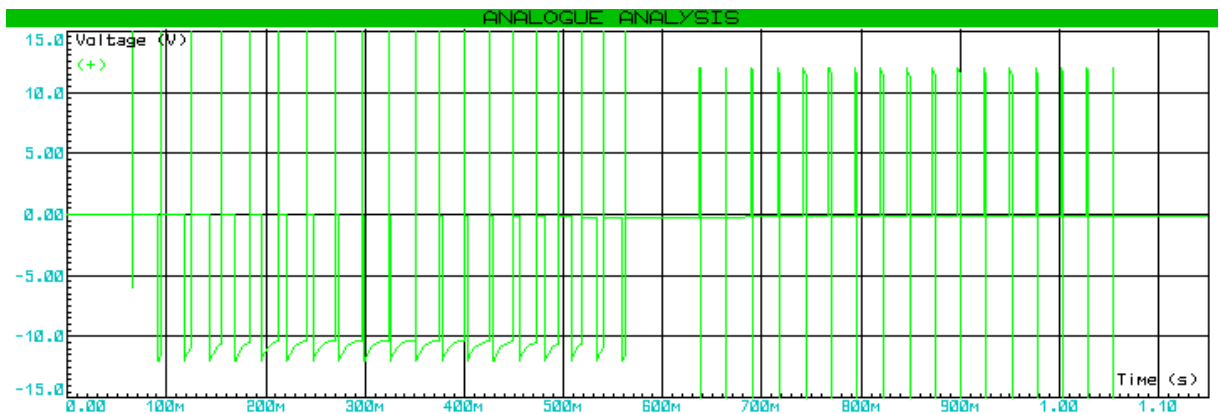

Figure 3.8. L298P output voltage when $V_{Left}$ is inputted

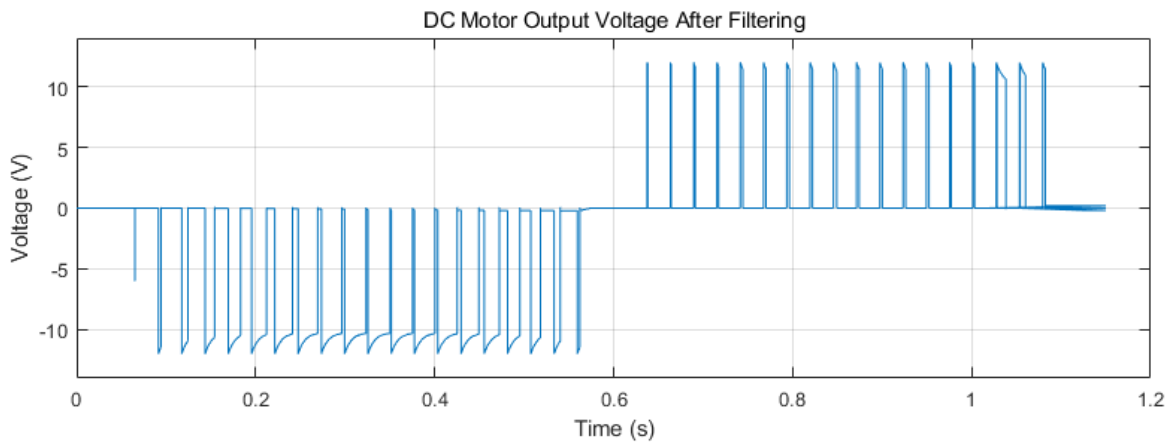

Figure 3.9. L298P output voltage when $V_{Left}$ is inputted after filtering

The maximum duty ratio $V_{Right}$ at negative voltage is smaller comparing with duty ratio at positive voltage. which meets with the requirement described in section 3.1 on the numerical expression of $V_{Right}$ that $V_{A1}$ should smaller than $V_{A2}$.

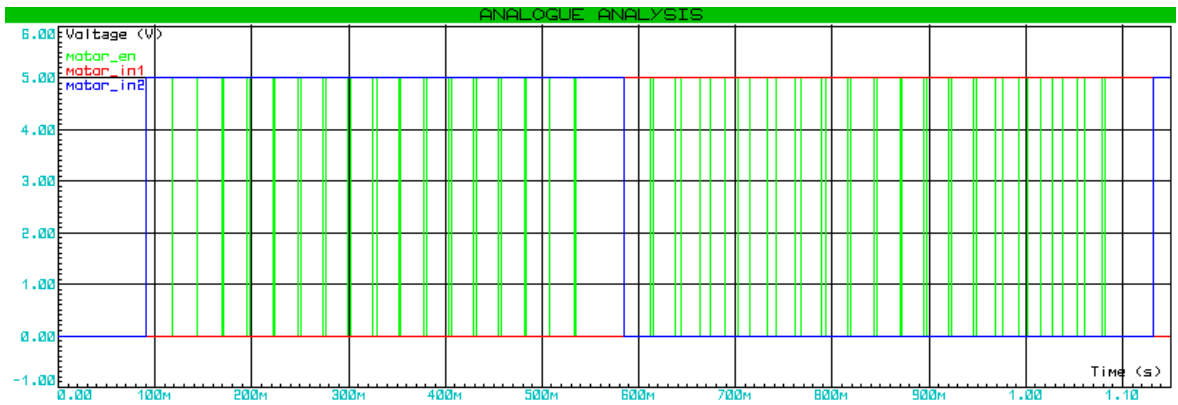

Figure 3.10. $V_{Right}$ control input signals to L298P

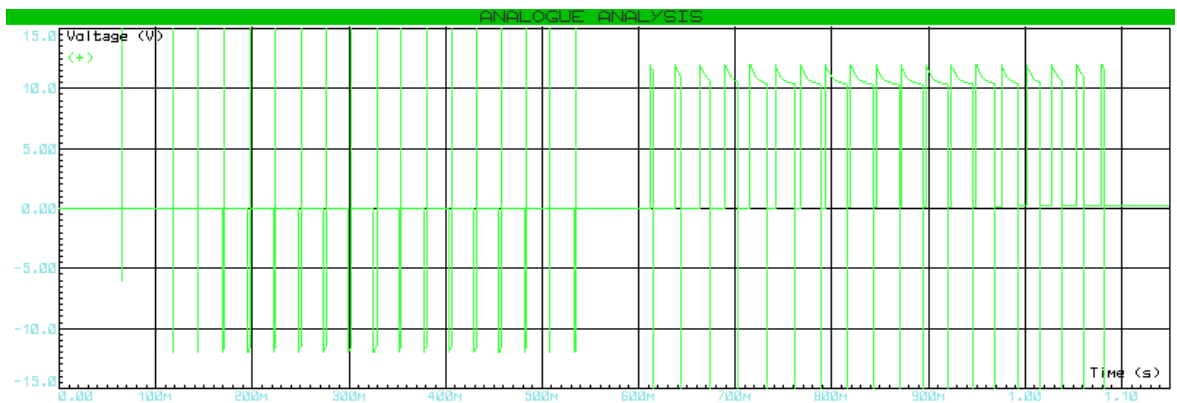

Figure 3.11. L298P output voltage when $V_{Right}$ is inputted

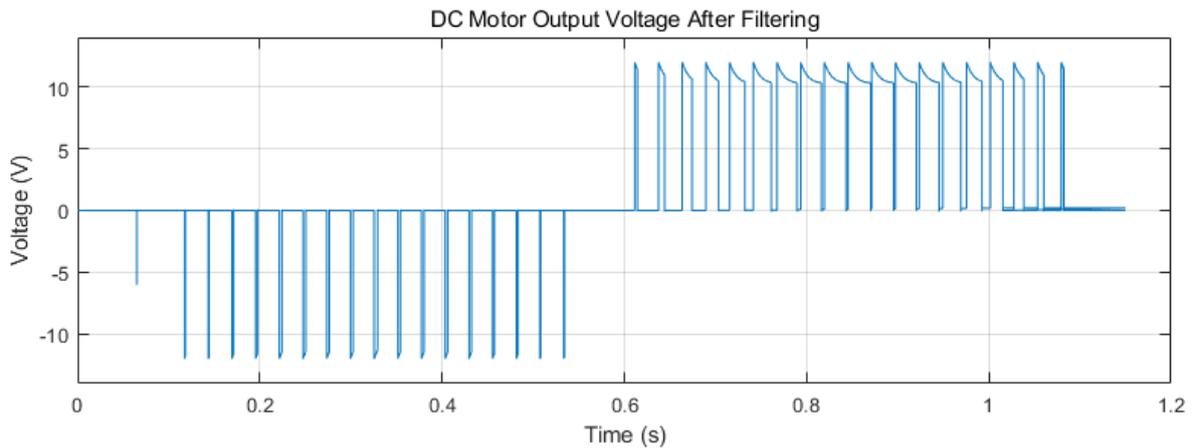

Figure 3.12. L298P output voltage when $V_{Right}$ is inputted after filtering

The output voltage from another L298P module to control the stepper motor is shown in the Figure 3.13 below. When setting the delay time between each step into 250 microseconds in the Arduino IDE programme, it can be measured from the graph that the actual time interval between each step is approximately five milliseconds. The step angle calculated in equation 2.1 is 1.8 degree, which means the linear speed of the metal move along the lead screw in the program can be derived by equation (2.2):

$$0.002 \times \frac{1.8° \times \frac{1}{0.005}}{360} = 0.002 \ m/s = 2 \ mm/s \tag{3.4}$$

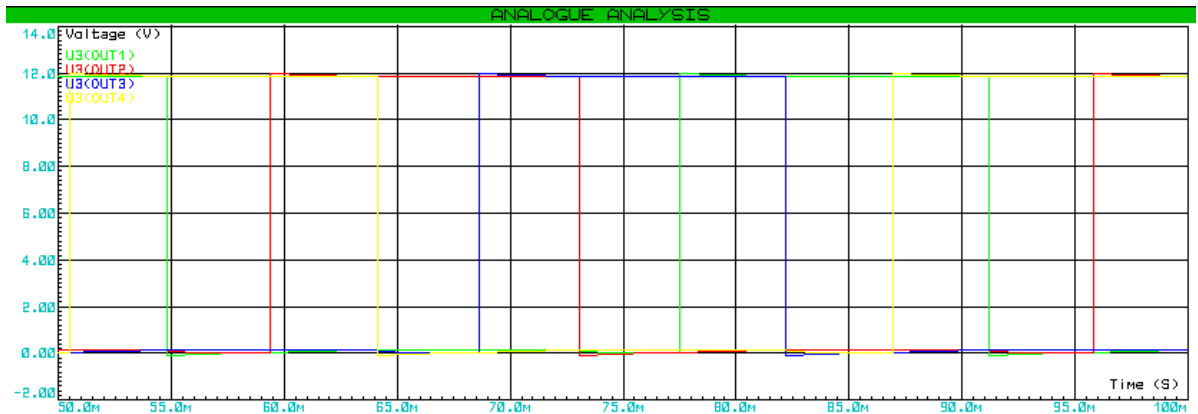

Figure 3.13. Stepper motor control signals output from L298P

The output voltage signals for controlling 2 servo motors are shown in the Figure 3.14 below. The total time period of the one cycle shown in the graph is 20 milliseconds and the pulse takes 1 millisecond and 2 milliseconds, corresponding to 0 degree and 90 degrees position.

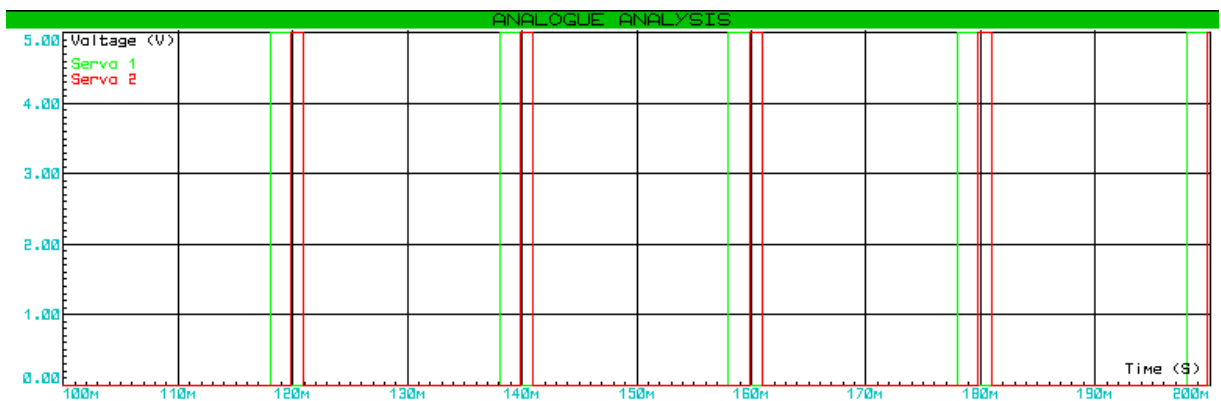

Figure 3.14. Servo motor control signals

### 3.3. FEM Simulation on the Soft Actuator

The prototyping of soft fluid actuators is time-consuming and complicated. As described in section 2.2.1, the mould needs to be manufactured by a stereolithography 3D printer to achieve high precision and avoid rough surfaces. After the printing, the mould needs to be washed with alcohol to remove the liquid resin and cured under UV light to increase the strength. The liquid silicon injection and removing bubbles will also take a significant amount of time. Generally, the manufacturing of a new version of soft fluid actuators from design will take at least a week, and the morphing of soft fluid actuators is difficult to directly predicted without testing, and thus simulation is effective to increase the efficiency of design and prototyping.

As a widely used method to analyse mathematical physics models, the Finite Element Method is useful to analyse the soft actuator's morphing under different pressure to validate and improve the design.

Simple linear shear-strain relationships can be used for the analysis of solid mechanics materials that withstand small strains. However, the analysis of soft robotics structures typically involves the highly flexible elastomers, in this case, silicone rubber. The deformation of these materials is usually larger than normal solid materials and is non-linear. Hyper-elastic solid mechanics models are needed to analyse the morphing of the soft actuator under different pressures.[22]

Two Parameters Mooney-Rivlin Model is chosen to simulate the soft actuator since it is suitable for moderate deformation. The model equation is given by:

$$W = C_1(I_1 - 3) + C_2(I_2 - 3) \tag{3.5}$$

$$I_1 = tr(C), \quad I_2 = \frac{1}{2}\left(tr(C)^2 - tr(C^2)\right) \tag{3.6}$$

where C is the right Cauchy-Green deformation tensor, and W is strain energy density.[8] Model parameters $C_1$ and $C_2$ for Ecoflex 30 silicone rubber is 48 kPa and -152 kPa, respectively. [23]

COMSOL Multiphysics is used to implement the FEM simulation by importing the soft actuator and apply equal distributed boundary loads on all internal faces within the cavity of the left and right actuator, respectively, to simulate the pressure difference between cavity and external environment.

The model of the robotic fish needs to be simplified to successfully generate mesh for simulation since the meshing algorithm is sensitive to the topological structure of the geometry. The convergence time could be significantly reduced with the simplified model as well. The simplified model is shown in Figure 3.15.

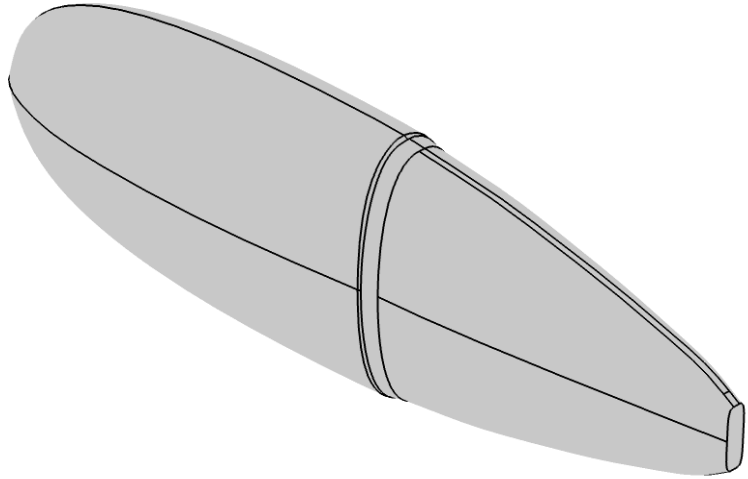

Figure 3.15. Simplified Model of the fish for FEM simulation

The model consists of 2 main components: rigid fish body and the soft actuator. The material of the fish body was set to ABS, and silicone was assigned to the soft actuator. The linear elastic material parameters and hyperelastic parameters used for simulation are shown in table 3.2 below.

Table 3.2. The FEM Simulation Parameters

| Parameter<br>Material | Young's modulus | Density | Poison's ratio | $C_1$ | $C_2$ |
|---|---|---|---|---|---|
| Silicone (Ecoflex 30) | N/A | N/A | N/A | 48 kPa | -152 kPa |
| ABS | 1.19e9 Pa | 1100 $kg/m^3$ | 0.29 | N/A | N/A |

Assuming the material is incompressible, by gradually increase and decrease the pressure until the simulation can no longer converge. The maximum boundary load that can converge is tested to be 44500 Pa and -8000 Pa. From the result, in order to achieve the fishtailing behaviour, the boundary load applied on the left and right actuator is given by functions

$$P_{left}(t) = \begin{cases} 44500 \times sin(0.125\pi t), & 2n \gg t > 2n+1 \\ 8000 \times sin(0.125\pi t), & 2n+1 \gg t > 2(n+1) \end{cases}, n = 1,2,3,4,5,\ldots,\infty \quad (3.6)$$

$$P_{right}(t) = \begin{cases} -8000 \times sin(0.125\pi t), & 2n \gg t > 2n+1 \\ -44500 \times sin(0.125\pi t), & 2n+1 \gg t > 2(n+1) \end{cases}, n = 1,2,3,4,5,\ldots,\infty \quad (3.7)$$

The functions of the pressure applied to the left and right actuator with time are plotted in MATLAB and shown in Figure 3.16 below.

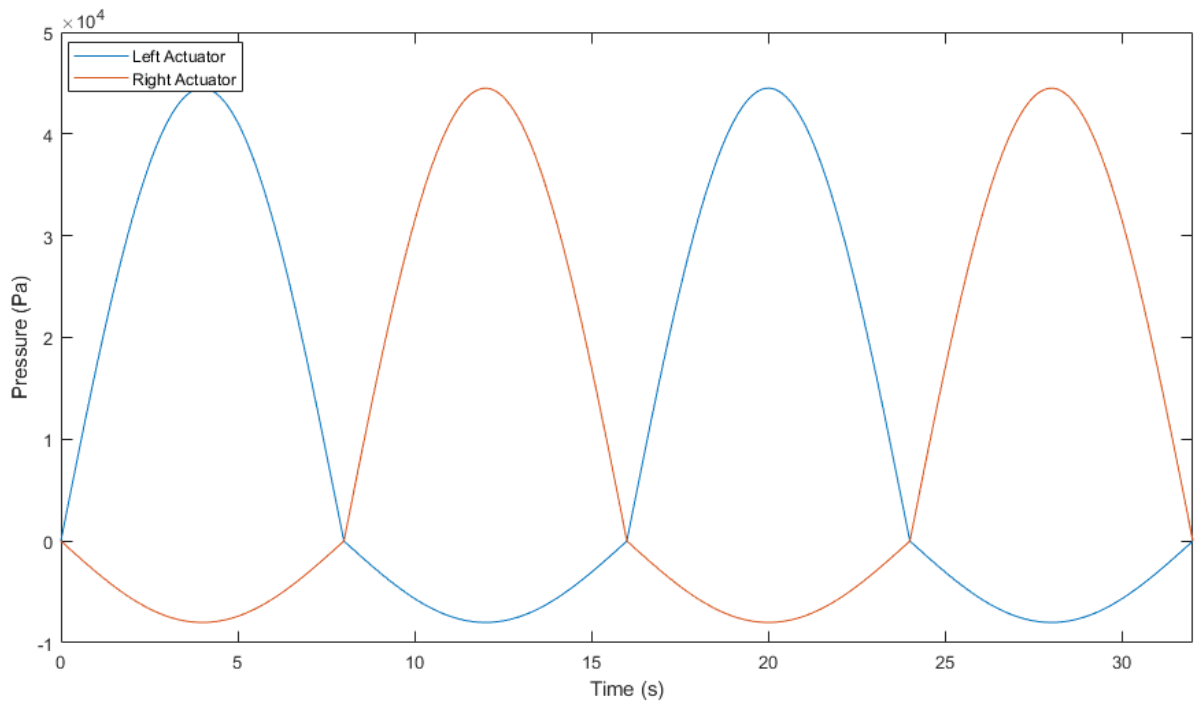

Figure 3.16. The plot of the pressure applied to the left and right soft actuator with time

By setting the study to time-dependent and setting the time range and interval to 16 s and 0.1 s, and running the simulation, the performance of the soft actuator can be simulated. However, it can be seen from the equations that the frequency is much lower than the actual fishtailing frequency. The reason is that if the frequency increases, the simulation calculation could not converge since higher frequency oscillation increases the instability in the system. It is possible to achieve a higher frequency in reality.

### 3.4. FEM Simulation Results

The time-dependent simulation results are shown in Figure 3.17. The time interval of each plot is 2 seconds. It can be seen from the results that the fishtailing actuation could be achieved by pressurising and depressurising the left and right actuator, which proves the design concept of the actuator. The plots show that the displacement at the end of the simplified model is approximately 40 mm in one direction, and because the tailing fins was excluded from the model, the estimated displacement at the end of the tail fins could reach approximately 120 mm in total in the simulation.

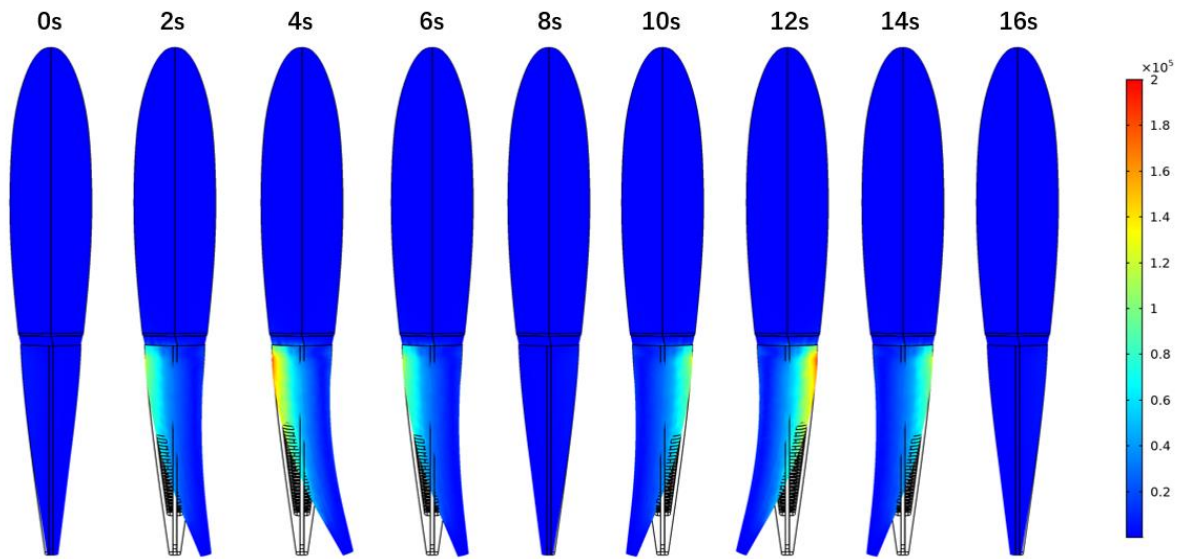

Figure 3.17. FEM simulation result

Two Parameters Mooney-Rivlin Model could no longer converge if further increase the pressure difference between the actuator cavities and external environment because the model is suitable to simulate moderate deformation. To observe larger deformation, the model could be simulated with Yeoh Model and Ogden Model, it was found that for Yeoh Model and Ogden Model, the deformation at the same pressure is much larger comparing with Mooney-Rivlin Model.

### 3.5. CFD Simulation on the Shell and Pectoral Fins

Although biomimetic streamlined shape design could significantly increase the swimming efficiency of the robotic fish in the fluid, the design of the robotic fish still faces the constrain of the volume of electronic components, and the electrical sensors and ports need to be placed on the shell, like the pyramid-shaped water sensor and electronic ports to charge the battery and upload the program. Since the object underwater could face 800 times larger drag comparing with the same object in the air [24], assuming the same velocity, an analysis of the hydrodynamic shape of the robotic fish is important to improve the swimming efficiency.

The pectoral fins connecting with the servo motor is designed to respond to the change in the turbulence and provide dynamic lift force as well as collaborate with the fishtailing actuator to turn the fish. The hydrodynamic simulation could validate the function and performance of the pectoral fins.

Computational Fluid Dynamics (CFD) is a useful tool to analyse the hydrodynamic shape and structure of the design. Solidworks Flow Simulation add-in allows the user with little basis in CFD simulation to easily set up simulation environments by following the steps in the Wizard. The simulation parameters are shown in Table 3.3 below. After the simulation is finished, the 3D flow trajectories can be visualised in the simulation result.

Table 3.3. The CFD Simulation Parameters

| Pressure | Temperature | Velocity in x-direction | Velocity in y-direction | Velocity in z-direction | Turbulence intensity | Turbulence length |
|---|---|---|---|---|---|---|
| 101325 Pa | 293.2 K | 0.5 m/s | 0 m/s | 0 m/s | 0.1% | 0.0014977 m |

## 3.6. CFD Simulation Results

Three models were simulated in the Solidworks Flow Simulation. Each has different pectoral fins angles (0 degrees, 30 degrees, and -30 degrees). The 3D flow trajectories of the simulation results are shown in Figure 3.18, 3.19, and 3.20 below.

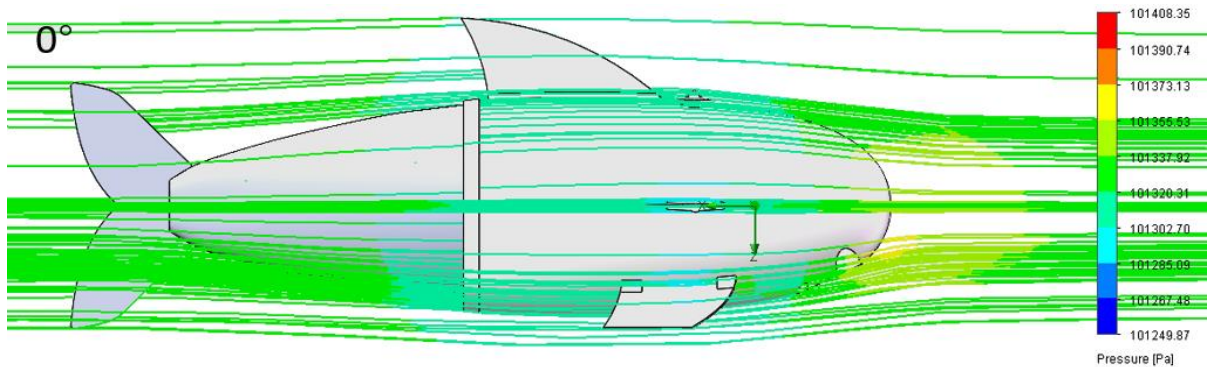

Figure 3.18. 3D flow trajectories of the robotic fish with 0-degree pectoral fins

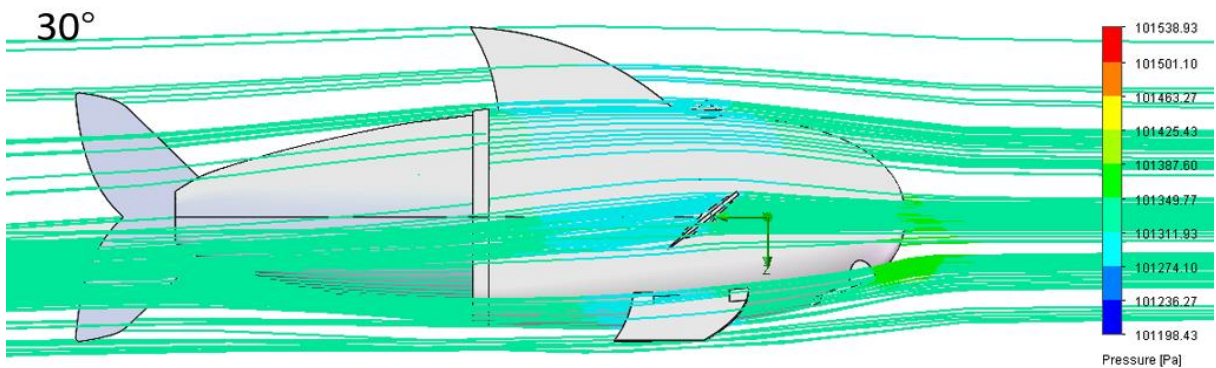

Figure 3.19. 3D flow trajectories of the robotic fish with 30-degree pectoral fins

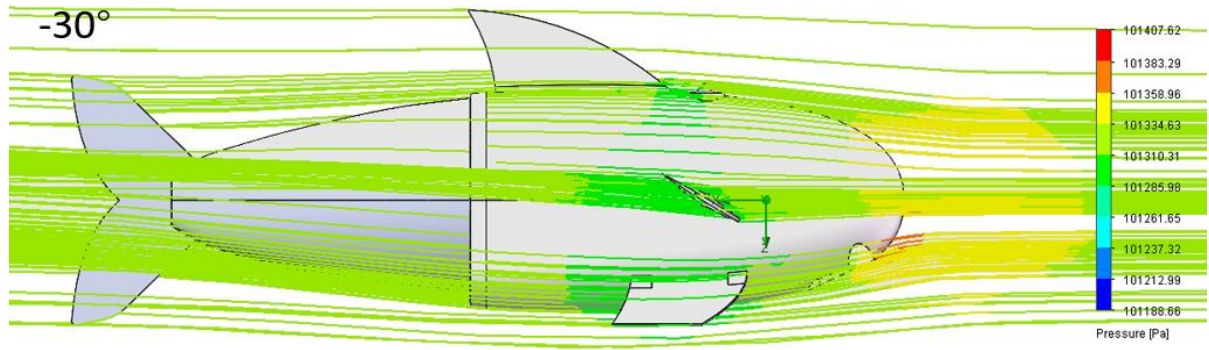

Figure 3.20. 3D flow trajectories of the robotic fish with -30-degree pectoral fins

It can be seen from the graph of the 3D flow and legend that the robotic fish has an overall hydrodynamically efficient streamlined shape. The simulation sets the pressure to 101.325 kPa, which is one standard atmosphere. When facing the 0.5 m/s head-on turbulence, the maximum pressure on the flow trajectories is only 101.408 kPa when the pectoral fins are parallel to the turbulence, slightly higher than the environment pressure. When the pectoral fins are in 30 degrees and -30 degrees angle with the turbulence, the maximum pressure on the flow trajectories is 101.538 kPa and 101.407 kPa, respectively.

The results also show that the change in the angle between pectoral fins and turbulence could change the direction of flow trajectories to control the dive depth and change direction. It is also worth paying attention that the simulation shows the maximum pressure generated near the electronic ports cover, which means the design of the electronic ports cover is hydrodynamically inefficient. Thus, further improvement can be made to the design of electronic ports to increase swimming efficiency.

## 4. Conclusion

### 4.1. Summary

Generally, the final project report covers the detailed mechanical and electronic design of a soft robotic fish and simulations on the subsystem performance. The design is based on the analysis of the pro and cons of existing designs in the literature. It has the following features to ensure the performance during the applications like aquatic creatures' study:

Firstly, the soft silicone-made actuator with embedded flexible thin-film curvature sensor enables closed-loop control on the actuator to improve the precision of the swimming trajectory.

Secondly, the redesigned gear pump could pressurise and depressurise the soft actuator by delivering strong, inversible and pulse-free water current. Thirdly, the IMU on the robotic fish could provide reliable measurement data on the pitch and roll angle to allow closed-loop control on fish position by adjusting the balancing control unit and artificial pectoral fins. Fourthly, the compact application board design places the electronic components on both sides of the PCB and connects to the Arduino Micro through two 17-pin headers. Such design significantly reduces the width of the fish body and increases the hydrodynamic efficiency of the fish. Finally, the 12 V battery pack consists of 3 protected lithium-ion rechargeable cells, and the total 3400 mAh capacity ensures the robot's operation time underwater. Built-in protection circuit reduces the risk of burning the application board in the event of water leakage.

A simplified control algorithm is designed to test the performance of the application board. The results meet the expectation and validate the functionality of the partial electronic designs. FEM simulation is used to prove the design concept and visualise the performance of the soft actuator. Although the simulation successfully proves the design concept, the results show that applying different numerical models could result in quite different morphing under the same condition. Thus, physical testing to determine the best fit numerical model is necessary. The hydrodynamic efficiency is checked by conducting CFD simulation, and the overall result shows excellent hydrodynamic efficiency, a potential improvement that can be made is also discovered through CFD simulation.

### 4.2.    Future Work

Due to the restriction of the project time and lab access, there are a couple of simulations and physical testing that can be done in the future to improve the design.

The manufacturer of Spectra Symbol Flex Sensor, the flexible curvature sensors placed in between the left and right actuators, does not provide a detailed and reliable resistance-curvature relationship, and according to the feedback from the people that purchased the sensor, the output signal of the sensor is noisy. Physical testing needs to be conducted to determine the resistance-curvature relationship, sensitivity, and noise of the sensor. If the result is unsatisfactory, the eGain sensor in Yu-Hsiang's design mentioned in section 1.2 can be an alternative solution.

The centre of gravity should lie on the major axis of the robotic fish since all of the components, except for some of the electronic component on PCB, which can be ignored, are placed symmetrically about the medial axis. The balancing control unit can adjust the centre of gravity along the medial axis to change the pitch angle of the robotic fish. However, it is difficult to determine the exact position of the centre of gravity through Solidworks. Theoretically, it can be achieved by assigning material to all the components in the assembly. Solidworks can automatically calculate the centre of gravity position. However, the 3D drawing of some of the components, like the battery and Stepper motor, only shows the model's appearance while ignoring the inside structure, which makes the exact centre of gravity hard to be determined at this stage. The weight of the metal cubic on the balancing control unit, metal cubic balanced position, and travel distance of the metal needs to be calculated and tested at the physical manufacturing stage. Furthermore, the buoyancy of the robotic fish needs to be manually adjusted to swimming at a deeper depth. How to develop the mechanism to automatically change the buoyancy according to the depth of dive while not affecting the position of the centre of gravity need to be considered in future design.

The simplified control algorithm is aimed to test the performance of the application board and microcontroller. The simulations on closed-loop control of fishtailing actuation and balance could be achieved through MATLAB Simulink if the sensor can be physically tested. Similar work has been done in Yu-Hsiang's paper. The simulation is crucial for developing a full control algorithm of the robotic fish.

The current design does not consider the remote wired or wireless control of the robotic fish. The radio signal could travel a long distance through the air but attenuate rapidly in water. Communication underwater is normally done through the acoustic signal. An acoustic communication modem can be added to the robotic fish to enable remote control of the robotic fish. However, the distance for valid acoustic communication is very short. Another solution is adding a camera at the reserved space at the front of the robotic fish, autonomous swimming can be achieved through image recognition algorithms, but more sensors like pressure sensor to measure the depth of the fish are needed to achieve autonomous swimming.

# 6. Appendices

## 6.1. Comparison for the Existing Designs

| Robotic Fish Design | Actuation | Trajectory Control | Advantage | Disadvantage |
|---|---|---|---|---|
| 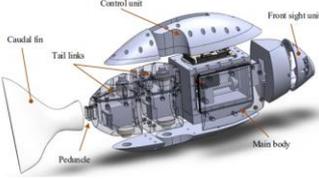<br>Figure 6.1. Virtual view of Mustafa's design [4] | Rigid Two-link tail mechanism driven by servo motor | CoG control unit use lead screw mechanism | 3D swimming trajectory<br><br>Effective depth and position control | Large volume and hydrodynamically inefficient<br><br>High power consumption |
| 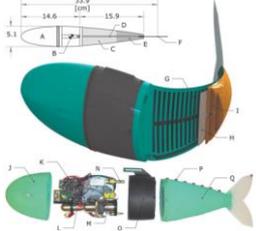<br>Figure 6.2. The anatomy and explosive view of Andrew's design [5] | Pneumatic fluidic elastomer actuators powered by a $CO_2$ cartridge | 2D plane trajectory swimming with asymmetrical fishtailing | Escape Maneuvers<br><br>Fast fishtailing frequency<br><br>Power efficient | Cannot perform depth control<br><br>Operation time limited by gar cartridge volume |
| 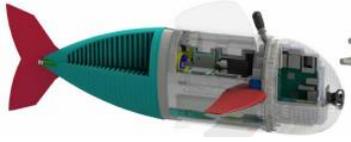<br>Figure 6.3. The anatomy view of Robert's design [6] | Hydraulic fluidic elastomer actuators powered by a gear pump | BCU with symmetrical piston structure and adjustable dive planes | 3D swimming trajectory<br><br>Acoustic communication modem<br><br>Power efficient | Require manual adjustment on weight when reach certain depth |
| 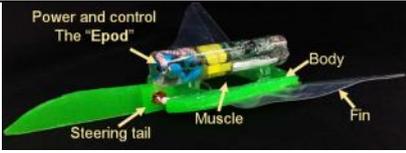<br>Figure 6.4. The image of Tiefeng's design [7] | Dielectric elastomers soft actuator | Cannot control swimming trajectory | Extremely power efficient<br><br>Low operating noise | Actuator needs large voltage boost to operate<br><br>Cannot control swimming trajectory |

## 6.2. Full Schematic Diagram of Application Board

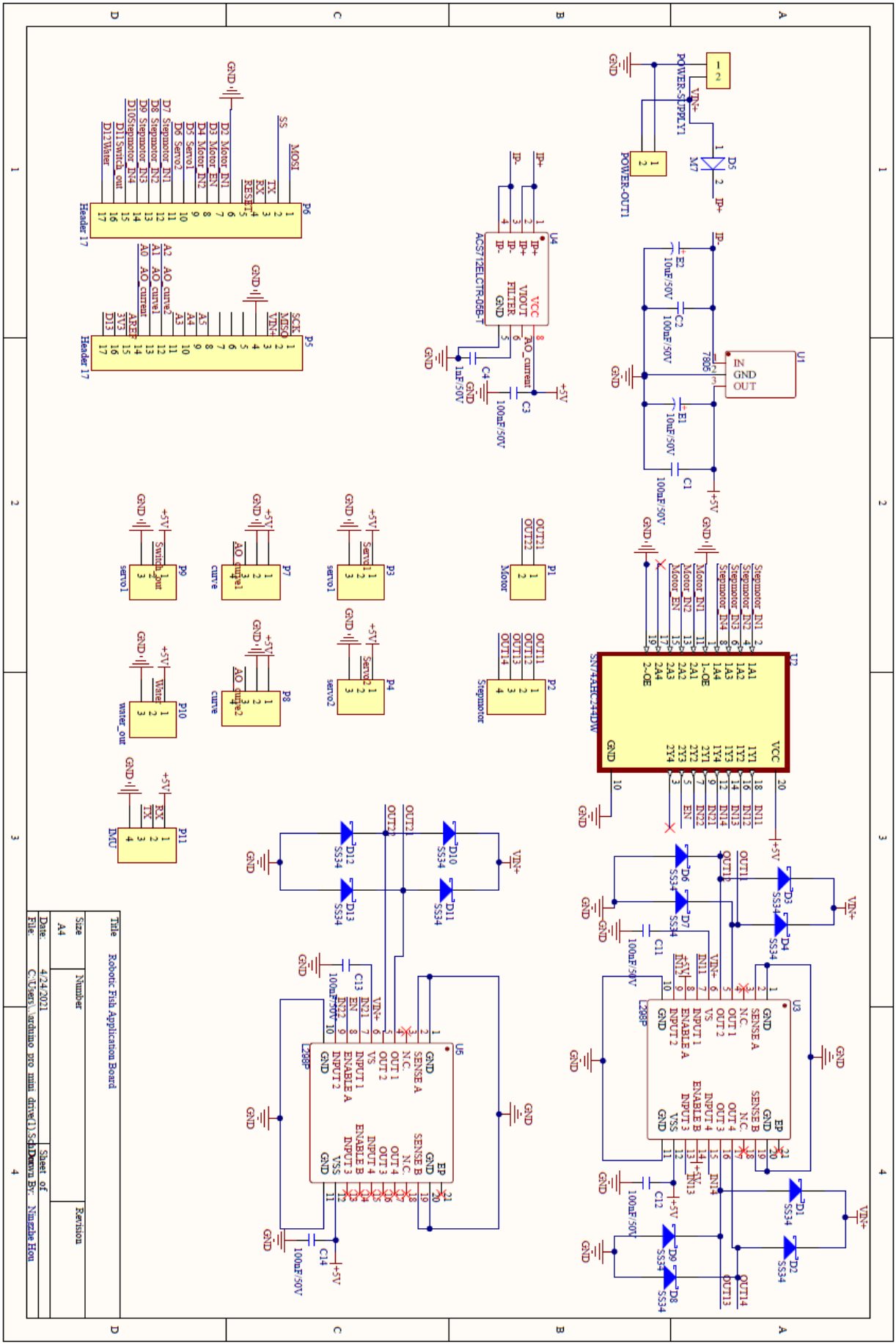

## 6.3. Application Board Wring Diagram and Top and Bottom View of PCB

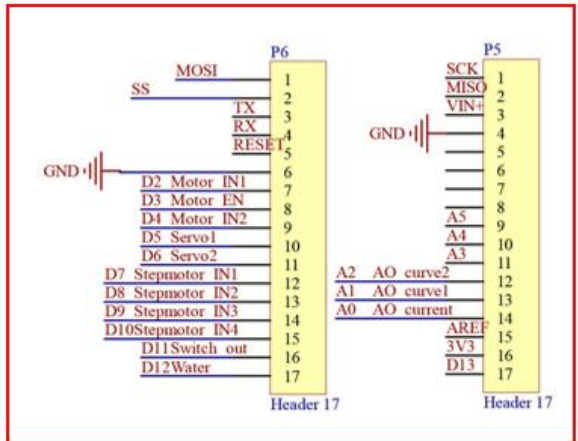
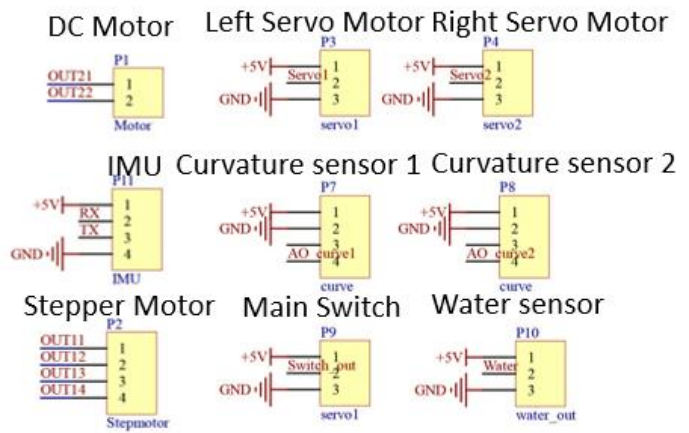
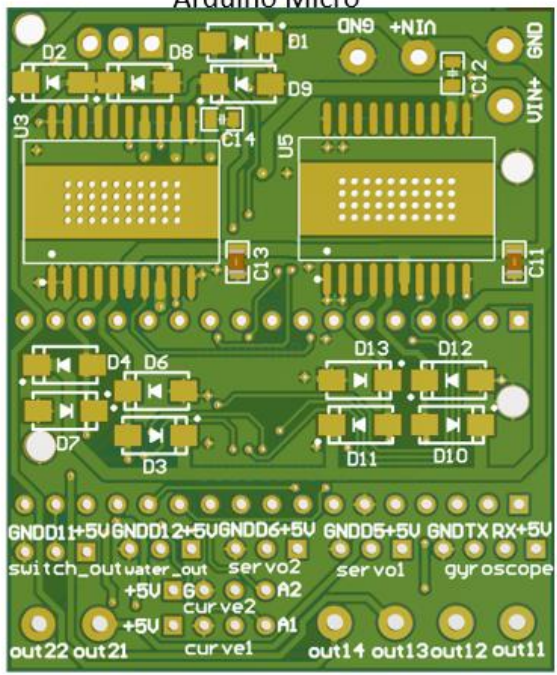
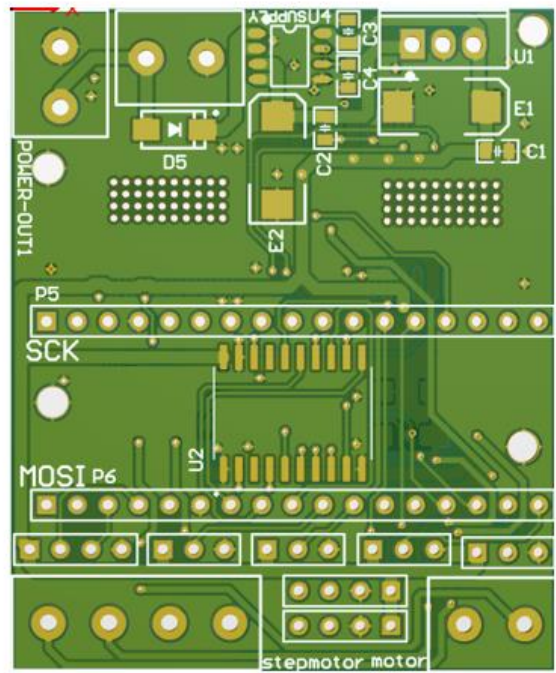

## 6.4. The Arduino IDE Source Code of Simplified Control Algorithms for Simulation

```
#include <LiquidCrystal.h>
#include <math.h>          //sin
#include<MsTimer2.h>   //timer used to genrate pwm
#include<Servo.h>   //servo motor
#include<AccelStepper.h> //stepper motor
Servo servo1;
Servo servo2;
#define pi 3.1415926
#define step1 8
#define step2 10
#define step3 2
#define step4 4
#define motor_in1 6
#define motor_in2 7
#define motor_en 9
#define key1 A0
#define key2 A1
```

```
#define key3 11
#define key4 12
#define key5 13
#define stepperSpeed 5
#define STEPS 648
AccelStepper stepper(AccelStepper::FULL4WIRE, 8, 10, 2, 4);
int t=0;
int a=0;//define time variable
int x=0;
int key_value=2;//mode switch, 1, 2, 3 means straight,left,right
int model=1;
int stepcountor=0;//recording the steps of stepper motor
unsigned char val1,val2;     //define sin wave, val1 has amplitude of 1, val2 has
amplidue of 0.5

void setup() {
pinMode(0, OUTPUT);
pinMode(1, OUTPUT);
pinMode(2, OUTPUT);
pinMode(3, OUTPUT);
pinMode(4, OUTPUT);
pinMode(5, OUTPUT);
pinMode(6, OUTPUT);
pinMode(7, OUTPUT);
pinMode(8, OUTPUT);
pinMode(9, OUTPUT);
pinMode(10, OUTPUT);
pinMode(11, INPUT);
pinMode(12, INPUT);
pinMode(13, INPUT);
servo1.attach(3);
servo2.attach(5);
  MsTimer2::set(1, flash);      // generate interrupt every 1 ms
  MsTimer2::start();             //timer start
  stepper.setMaxSpeed(100);
    stepper.moveTo(0);
}

void flash()                          //interupt function for genrrating pwm
{
    t++;
    x++;
    if(x>800)
    x=0;
    if(t>25){
    t=0;
    a++;
    if(a>40)
    a=0;
    switch(model)
    {
    case 1:
    if(a<20)
    {
       digitalWrite(motor_in1,LOW);
       digitalWrite(motor_in2,HIGH);
```

```
      val1=25.0*sin(pi*a/20.0);
    }
    else
    {
      digitalWrite(motor_in1,HIGH);
      digitalWrite(motor_in2,LOW);
      val1=25.0*sin(pi*(a-20)/20.0);
    }
    break;
    case 2:
    if(a<20)
    {
      digitalWrite(motor_in1,LOW);
      digitalWrite(motor_in2,HIGH);
    val1=25.0*sin(pi*a/20.0);
    }
    else
    {
      digitalWrite(motor_in1,HIGH);
      digitalWrite(motor_in2,LOW);
    val1=4.0*sin(pi*(a-20)/20.0);
    }
    break;
    case 3:
    if(a<20)
    {
      digitalWrite(motor_in1,LOW);
      digitalWrite(motor_in2,HIGH);
    val1=4.0*sin(pi*a/20.0);
    }
    else{

      digitalWrite(motor_in1,HIGH);
      digitalWrite(motor_in2,LOW);
    val1=25.0*sin(pi*(a-20)/20.0);
    }
    break;
    default:
    val1=25.0*sin(pi*a/40.0);
    }
    }
    if(t<val1)
    digitalWrite(motor_en,HIGH);
    else
    digitalWrite(motor_en,LOW);

}

void key_scan(void)
{

  if(digitalRead(key1)==0)
   {
    delay(2);
    if(digitalRead(key1)==0)
    {
    key_value=1;
      stepper_reset(); //stepper return to 0 if while swimming straight
    }
```

```
  }

    if(digitalRead(key2)==0)
    {
    delay(2);
      if(digitalRead(key2)==0)
      {
      key_value=2;
       stepper_reset();//stepper return to 0 if while swimming left
      }
    }

    if(digitalRead(key3)==0)
    {
    delay(2);
      if(digitalRead(key3)==0)
      {
      key_value=3;
        stepper_reset();//stepper return to 0 if while swimming right
      }
    }

    if(digitalRead(key4)==0)
    {
    delay(2);
      if(digitalRead(key4)==0)
      {
      key_value=4;
        stepper_up();//stepper spinning clockwise 3 rev
      }
    }

    if(digitalRead(key5)==0)
    {
    delay(2);
      if(digitalRead(key5)==0)
      {
      key_value=5;
        stepper_up();//stepper spinning counter clockwise 3 rev
      }
    }
}
void V_Straight(void)
{
  model=1;
}

void V_left(void)
{

  model=2;
}

void V_right(void)
{
    model=3;
}
```

```
void stepper_reset()
  {
  stepper.moveTo(0);
    stepper.run();
  }

  void stepper_up()
{
    stepper.moveTo(600); //3 rev
      stepper.run();

}
  void stepper_down()
{
    stepper.moveTo(-600); //3 rev
      stepper.run();
}

void drive(void)
{
  switch(key_value)
  {
  case 1://straight forward
  V_Straight();
  servo1.write(45);
  servo2.write(45);
  break;
  case 2://left turn
  V_left();
  servo1.write(75);
  servo2.write(15);
  break;
  case 3://right turn
  V_right();
  servo1.write(15);
  servo2.write(75);
  break;
  case 4://elevator up
  V_Straight();
  servo1.write(15);
  servo2.write(15);
  break;
  case 5://elevator down
  V_Straight();
  servo1.write(75);
  servo2.write(75);
  break;
  default:
  servo1.write(45);
  servo2.write(45);
  }
}
void loop() {
  key_scan();
  drive();

}
```

## 6.5. COVID-19 Impact Statement

As required by the department, this statement is attached in the Appendix to describe the impact of COVID-19 on the project and actions taken to minimise the project's negative impact.

Since this is a bespoke project designed in August, back then, the daily Covid-19 cases had been under control, and the lockdown restriction had been lifted across the UK. In the original planning, two parts were included in the project: virtual mechanical and electronic design and physical manufacturing.

However, the daily COVID-19 cases exponentially went up, and the whole country was in lockdown again since early October 2020.

Initially, with the help of the project supervisor, Dr Alexandru Stancu, several attempts were made to realise the physical manufacturing. He helped me with prototyping some of the 3D printed parts, and I reached out to a Swiss silicon rubber 3D printing start-up company for the manufacturing of soft actuator. But since the later school announcement that the campus facilities were going to remain closed, which means even the assembly of the components at the lab was impossible, the decision was made to replace all the physical manufacturing and testing with software simulation in early January.

Luckily, the results of the three simulations were satisfactory and sufficient to validate the performance of the robotic fish. In addition, Peking University Intelligent Biomimetic Design Lab are willing to help with the physical manufacturing of the robotic fish during the summer holiday of 2021.